\documentclass[]{aiaa-tc}

\usepackage[detect-all,per-mode=symbol]{siunitx}
\usepackage{booktabs}
\usepackage{amssymb}	
\usepackage{amsmath}	
\usepackage{commath}	

\usepackage{standalone}
\usepackage{pgfplots}
\pgfplotsset{compat=newest}
\usepgfplotslibrary{groupplots}        
\usepgfplotslibrary{polar}            

\usepackage[capitalize]{cleveref}

\newcommand{\dt}{\Delta t}
\newcommand{\xtt}{x_{t+\dt}}
\newcommand{\ztt}{z_{t+\dt}}
\newcommand{\btt}{b_{t+\dt}}
\newcommand{\xt}{x_{t}}
\newcommand{\bt}{b_{t}}
\newcommand{\ut}{u_{t}}
\newcommand{\zt}{z_{t}}
\newcommand{\seq}{\,{=}\,} 
\newcommand{\exz}{E_{z\in\mathcal{Z}}}
\newcommand{\thetae}{\theta^{\mathrm{e}}}
\newcommand{\xte}{x_t^{\mathrm{e}}}
\newcommand{\thetan}{\theta^{\mathrm{n}}}
\newcommand{\xtn}{x_t^{\mathrm{n}}}
\newcommand{\se}{\textrm{SE}(2)}

\pgfplotsset{every axis legend/.append style={legend cell align=left}}

\title{Efficient and Low-cost Localization of Radio Signals with a Multirotor UAV}

 \author{
  Louis Dressel%
    \thanks{PhD Candidate, Aeronautics and Astronautics, Stanford University, AIAA Student Member.}
  \ and Mykel J. Kochenderfer\thanks{Assistant Professor, Aeronautics and Astronautics, Stanford University, AIAA Associate Fellow.}\\ 
  {\normalsize\itshape
   Stanford University, Stanford, CA 94305}
 }

 \AIAApapernumber{YEAR-NUMBER}
 \AIAAconference{Conference Name, Date, and Location}
 \AIAAcopyright{\AIAAcopyrightD{YEAR}}

\begin{document}

\maketitle

\begin{abstract}
Localizing radio frequency (RF) sources with an unmanned aerial vehicle (UAV) has many important applications.
As a result, UAV-based localization has been the focus of much research.
However, previous approaches rely heavily on custom electronics and specialized knowledge, are not robust and require extensive calibration, or are inefficient with measurements and waste energy on a battery-constrained platform.
In this work, we present a system based on a multirotor UAV that addresses these shortcomings.
Our system measures signal strength received by two antennas to update a probability distribution over possible transmitter locations.
An information-theoretic controller is used to direct the UAV's search.
Signal strength is measured with low-cost, commercial-off-the-shelf components.
We demonstrate our system using three transmitters: a continuous signal in the UHF band, a wildlife collar pulsing in the VHF band, and a cell phone making a voice call over LTE.
Our system significantly outperforms previous methods, localizing the RF source in the same time it takes previous methods to make a single measurement.
\end{abstract}

\section{Introduction}
In robotic localization of radio frequency (RF) sources, a robot searches for the source of some RF signal.
The robot carries electronics and one or more antennas to analyze the received RF signal.
These RF measurements are usually distilled into range or bearing estimates to the RF transmitter.
The estimates are combined with knowledge of the robot's own position to update a belief, or probability distribution, over possible source locations.
The robot then reasons over the belief to select its next action, with the aim of minimizing time or energy spent searching.
The localization is considered complete when the belief is sufficiently concentrated in a small area, indicating a likely position for the RF source.

Fast localization of RF sources is important to many applications. For example, quickly locating RF sources like distress beacons can be critical in directing search and rescue teams and saving lives\cite{Hoffmann2006}. Similarly, quickly locating sources of GPS interference, such as GPS jammers, can be important in ensuring the integrity of air traffic systems\cite{Geyer1999}. Instruments at Newark Liberty International Airport experienced interference when drivers using ``personal privacy devices"---small GPS jammers---drove on the nearby New Jersey Turnpike\cite{gnss2}.
The Federal Aviation Administration investigated the use of a small, manned aircraft for localizing interference\cite{Geyer1997}, but manned solutions are expensive.

Even when safety of life is not threatened, efficient localization of RF sources is important.
For example, localizing radio-tagged wildlife is critical to understanding behavioral patterns of animals and can aid conservation efforts.
Unfortunately, current localization methods are time and resource intensive\cite{cliff2015}.
These methods rely on a human traversing terrain---often uninhabited and rough---and estimating bearing to a radio-tag by manually rotating a directional antenna\cite{posch2009}.
Using a robot could reduce the burden on researchers, allowing them to study more animals at once.
Further, a \textit{flying} robot, or unmanned aerial vehicle (UAV), is unencumbered by rough terrain; the value of aerially bypassing tough terrain has long been recognized by ecologists\cite{Seddon2004}.
Flying also reduces reflections from obstacles on the ground\cite{korner2010}.

Due to its many applications, localizing RF sources with UAVs has drawn significant attention.
For example, UAVs have been used to localize WiFi routers\cite{Perkins2015}, ZigBee radios\cite{Venkateswaran2013,isaacs2014}, GPS jammers\cite{Perkins2016}, and wildlife radio-tags\cite{cliff2015}.
However, UAVs have some drawbacks: they are less stable than ground robots and have limited payloads and flight times.
In an effort to reduce weight, some methods resort to custom circuity\cite{cliff2015}, but this can be unattractive if we want researchers of different fields to use and maintain the system, or if they want to augment their system to cover new frequencies.
Some use COTS electronics that only work on a single frequency\cite{Venkateswaran2013,isaacs2014,Perkins2015}, limiting the range of applications.
The most serious drawback in prior work is that most methods---including all examples cited in this paragraph---require the UAV to make a full rotation for a single bearing measurement.
Each rotation can take up to \SI{45}{\second}, which is a significant amount of the typical 10--15 minute available flight time.
This might be the greatest bottleneck in improving the speed of aerial RF localization.

In this work, we present a system designed to address these flaws.
Our system is based on comparing the signal strength received by two antennas carried by a multirotor UAV---one mounted facing forward, the other facing back. 
These antennas are lightweight, low-cost, and simple to build.
They feed into low-cost, commercial-off-the-shelf (COTS) dongles, which in turn feed into an onboard computer via USB.
Leveraging existing open-source software, we use these dongles to estimate the signal strength received by each antenna.
If the forward-facing antenna receives higher strength, the RF source is deemed to be in front of the UAV and vice versa.
Combining these measurements with the UAV's GPS position and heading, an estimate of the source's location is produced.
In the onboard computer, a greedy, information-theoretic planner uses this estimate and leverages the UAV's maneuverability to provide velocity commands.
We demonstrate this localization procedure in simulation and with real flight tests, localizing three RF sources.
One is an animal radio-tag in the VHF band, another is a ham radio emitting in the UHF band, and the last is a cell phone operating over LTE.

Our contributions are as follows:
\begin{enumerate}
	\item We present a novel, low-cost, UAV-based system for localizing general RF sources. The system is faster than most existing methods and requires no specialized knowledge of electronics.
	\item We analyze the performance of our system's sensing modality, which we call field-of-view sensing, to understand how its performance changes with noise and design parameters.
	\item We perform flight tests to validate our system against RF sources of different frequencies and operating modes.
		These tests include localization of a cell phone operating over LTE, which, to our knowledge, is a novel application.
\end{enumerate}

The paper begins with the background and related work that motivates our research.
\Cref{sec:overview} describes the system, and \Cref{sec:modality} describes our sensing modality.
Our localization algorithm is presented in \Cref{sec:algorithm}.
Simulation results are presented in \Cref{sec:simulations}, and flight test results are presented in \Cref{sec:experiments}.

\section{Related Work and Motivation}
\label{sec:background}

Early related work used UAVs to localize a magnetic rescue beacon~\cite{hoffmannthesis}.
However, the sensor modality leveraged unique properties of the beacon and tests occurred over a small $\SI{9}{\meter} \times \SI{9}{\meter}$ area.
A method for localizing wildlife radio-tags over larger distances with UAVs was proposed by Posh and Sukkarieh\cite{posch2009}; this work was built upon by K{\"o}rner et al. and ground tests were performed\cite{korner2010}.
Soriano et al. used a similar method and also carried out ground tests\cite{Soriano2009}.
These methods suggested using fixed-wing UAVs and modeling power expected from the given source and receiver locations. 
The measurement model relied on predicting the power that should be received by the UAV's antenna and comparing it to the true received power.

Although these methods performed reasonably well in simulation and limited ground testing, this measurement model is not robust or applicable to general RF source localization tasks.
For example, the measurement model requires knowledge of the RF source's transmitting power.
In some problems, such as GPS jammer localization, the type of transmitter is completely unknown to searchers.
Even when the transmitter specifications are known, signal strength is affected by many unmodeled factors.
In wildlife radio-tag localization, the tagged animal might transition behind vegetation or terrain, greatly reducing received power.
In addition, the radio-tag's antenna---even if nominally omnidirectional---likely has \textit{some} directionality to it, which can affect received power depending on the animal's pose.
In recent work, Bayram et al. used the difference between successive strength measurements to perform gradient ascent on signal strength.\cite{bayram2016}
This method obviates the need to know transmitter strength, as long as it remains constant during localization.
However, the method fails to address factors that might change between successive measurements, like animal pose or terrain occlusions.

The difficulty of using signal strength measurements makes bearing an attractive sensing modality for general RF localization.
A common way to estimate bearing to an RF source is to use beamsteering with an antenna array, which has the benefit of quickly estimating bearing.
However, beamsteering can be heavy, complex, and prone to calibration errors\cite{hood2011}.

As a result, small robots often estimate bearing by rotating a directional antenna~\cite{Graefenstein2009,hood2011}.
If the timescale of a rotation is significantly less than the timescales of factors affecting signal strength, strength measurements from the rotation can be used to estimate bearing to the RF source.
There are still factors that affect all measurements from the same rotation equally, like the unknown distance to the transmitter.
However, these can be normalized out.
The simplest normalization technique sets the bearing estimate to the heading at which the highest strength was recorded, negating factors affecting all measurements.
Another option is to subtract the mean from all measurements and divide by their standard deviation, allowing for cross-correlation with the antenna's reference gain pattern~\cite{Graefenstein2009}.

The rise of maneuverable UAVs has popularized the antenna rotation strategy because they do not need extra actuators to rotate an antenna.
Venkateswaran et al. exploit a vehicle that constantly rotates to keep itself airborne to measure bearing during the rotations.\cite{Venkateswaran2013}
Isaacs et al. extended this work to a multirotor UAV that constantly rotates.\cite{isaacs2014}
However, constantly rotating a multirotor UAV complicates any control strategies that might be used.
Instead, most recent work relies on rotating a multirotor UAV once, moving to a new location, and rotating for a new measurement~\cite{cliff2015,Perkins2015,Perkins2016,vonehr2016}.

Rotating a multirotor UAV to estimate bearing has two main drawbacks.
The first is speed.
Rotations are reported to take about \SI{25}{\second}~\cite{Perkins2015,Perkins2016}, \SI{40}{\second}~\cite{vonehr2016}, or even \SI{45}{\second}~\cite{cliff2015}.
Small UAVs typically have battery lives of 10--15 minutes, so a \SI{45}{\second} rotation could amount to nearly 8\% of available flight time.
Spending so much time for a single measurement limits the number of measurements that can be made.
It also slows down localization, which needs at least a few bearing estimates for a decent source position estimate.

The second drawback of rotate-for-bearing is that it assumes transmitter strength and the time-varying factors we identified earlier remain constant during each rotation.
However, it is possible that an RF source changes its transmit power with time---it could even be a strategy to avoid detection by rotate-for-bearing UAVs.
Further, factors like transmitter antenna directionality and orientation might change with timescales smaller than the duration of each rotation.
These factors could affect each measurement from the rotation differently---then a comparison between the measurements would not yield a good bearing estimate.

In the preceding paragraphs, we have briefly covered the evolution of RF localization with small UAVs.
We show that the chief drawback of current methods is the need to perform a complete rotation to make bearing measurements.
There are other drawbacks; most previous work only works on a single frequency and uses complex electronic circuitry.

\section{System Description}
\label{sec:overview}
In this section, we describe a system designed to overcome the limitations outlined at the end of the last section.
The core idea is attaching two directional antennas to a multirotor UAV and measuring received strength simultaneously.
If the front-facing antenna measures higher strength, the RF source is assumed to lie in front of the UAV; if the rear antenna measures higher strength, the RF source is assumed to lie behind the UAV.
These pseudo-bearing measurements provide less information than a bearing estimate but do not require the UAV to rotate in place. 
Because we compare measurements made simultaneously from both antennas, time-varying factors are normalized out.

\subsection{Airframe}
\begin{figure}
	\centering
	\includegraphics[width=3.1in]{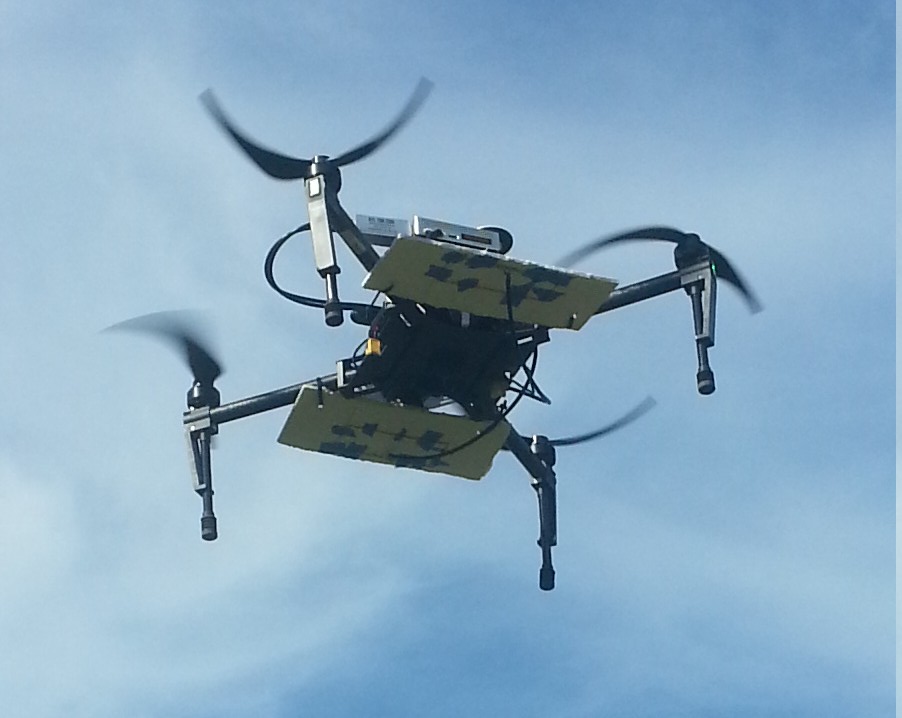}
	\caption{Our UAV in flight with \SI{782}{\mega\Hz} antennas mounted underneath.}
	\label{fig:matrice}
\end{figure}

We use a DJI Matrice 100 (M-100) quadcopter, which DJI markets as a stable airframe for developers of UAV applications.
In our work, the M-100 was both stable and easy to work with.
Including its battery, the M-100 is~\SI{2.4}{\kilogram} and has a max takeoff capacity of \SI{3.4}{\kilogram}, allowing for a \SI{1}{\kilogram} payload.
When carrying a payload, the M-100 has a flight time of about 15 minutes.
The M-100's maximum speed is \SI{22}{\meter\per\second}.
The M-100 has a built-in flight controller that provides low-level commands to the motors, keeping itself stable.
A serial connection to the flight controller allows flight data to be queried.
Position and velocity commands can be provided to the flight controller over the same link.
In our system, an onboard computer provides velocity commands, which the flight controller executes while taking care of low-level motor inputs to keep the UAV stable.
Our fully-equipped M-100 can be seen in \Cref{fig:matrice}.

\subsection{Moxon Antennas}
The first components of our RF subsystem are the directional antennas.
Because the gain provided by directional antennas depends on their orientation with respect to an RF source, they are useful in estimating bearing to RF sources.
Gain is usually highest when the antenna points directly at the source, leading to higher strength measurements.
The directionality of an antenna measures how tightly gain is concentrated.
High directionality is often desired in bearing estimation, because the high gain achieved in the main lobe implies a bearing to the RF source with low ambiguity.

Unfortunately, directional antennas can be large---especially at low frequencies, because antennas generally scale inversely with frequency.
Further, directional antennas generally grow with their directionality.
For example, one way to concentrate the gain of a Yagi antenna is to add elements, yielding a longer, heavier antenna.
Mounting these antennas on small UAVs can be difficult because of their weight and the space they occupy.

In order to limit antenna size, we use antennas that are only slightly directional.
Specifically, we use Moxon antennas~\cite{moxon1993}, which are similar to Yagi antennas with only two elements.
Moxon antennas are popular in the amateur radio community because they are easy to build and mechanically robust. 
They have low directionality, with most of their gain concentrated in a wide main lobe.

\begin{figure}
	\centering
	\includegraphics[trim={2.5in 3in 2.7in 2.7in},clip,width=3.2in]{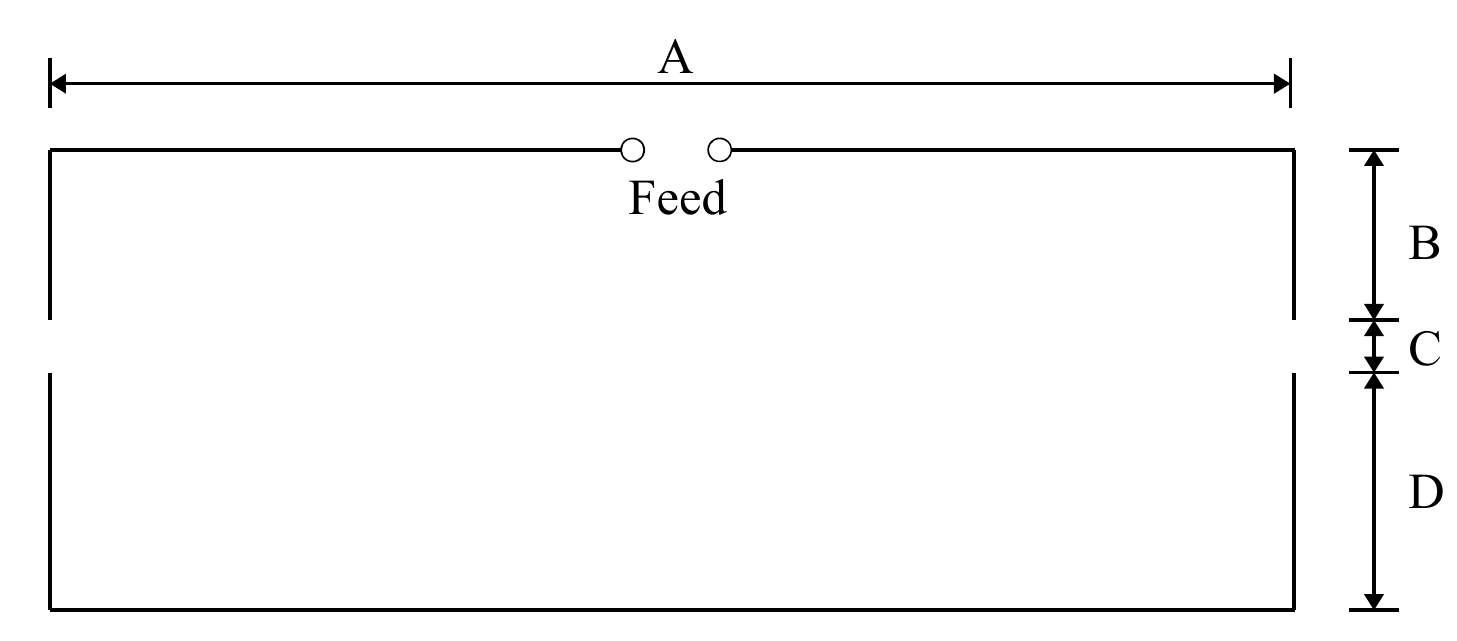}
	\caption{Top view of a basic Moxon antenna. Feed side points forward.}
	\label{fig:moxon}
\end{figure}

Design of these antennas requires no specialized electrical engineering knowledge.
Because Moxon antennas are popular with amateur radio enthusiasts, there exist many free Moxon design generators.
These tools interpolate between several well-modeled Moxon antennas for a range of frequencies and wire diameters.
We use one such Moxon generator~\cite{moxongenerator}, inputting only frequency and wire diameter.
\Cref{tab:antennas} shows the resulting antenna sizes for the frequencies of interest.
The Moxon generator also produces an input file for NEC-2, an antenna analysis tool developed for the U.S. Navy~\cite{burke1979}.
Now in the public domain, NEC-2 can be used to tweak Moxon designs further.
However, the Moxon generator designs performed well in practice and we deemed this step unnecessary.

\begin{table}
	\caption{Antenna sizes produced by Moxon generator~\cite{moxongenerator} for different frequencies and 14 AWG copper wire. Lengths A, B, C, and D correspond to those from \Cref{fig:moxon}. Mass includes coax cable.}
	\centering
    \begin{tabular}{@{} lccccc @{}}
    \toprule
	Frequency (MHz) & A (cm)& B (cm)& C (cm)& D (cm)& Mass (g)\\
    \midrule
	217.335 & 49.81 & 6.84 & 2.07 & 9.51 & 118 \\
	432.7 & 24.88 & 3.26 & 1.20 & 4.79 & 70\\
	782.0 & 13.71 & 1.71 & 0.75 & 2.66 & 66\\
    \bottomrule
    \end{tabular}
	\label{tab:antennas}
\end{table}

\begin{figure}
	\centering
	\includegraphics[width=3.4in,trim={0 6cm 0 0},clip]{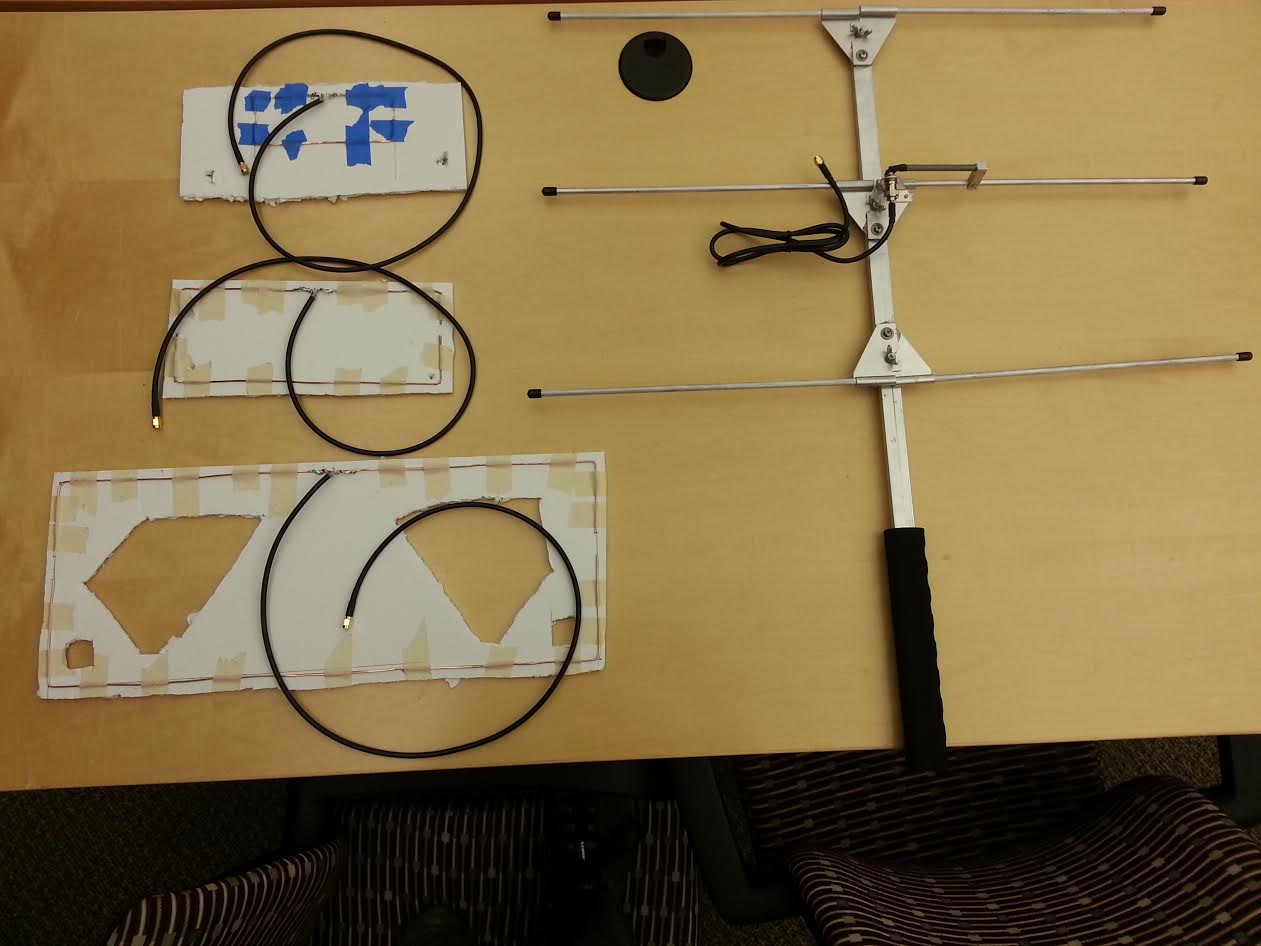}
	\caption{Our Moxon antennas on the left, from top to bottom: \SI{782}{\mega\Hz}, \SI{432.7}{\mega\Hz}, \SI{217.335}{\mega\Hz}. For size comparison, a commercially available \SI{217}{\mega\Hz} Yagi is on the right.}
	\label{fig:antennas}
\end{figure}

Construction is also trivial and requires no mechanical skill beyond rudimentary soldering.
The antenna dimensions are drawn on styrofoam board.
Copper wire---size 14 AWG in our case---is bent and cut to fit the dimensions.
The wire is taped to the board, and a thin cut is made in the upper wire for the feed.
A suitable length of RG-58 coaxial cable is selected (about half a meter) and an inch of the cable is stripped of its inner and outer insulation.
The inner conductor is soldered to one side of the copper feed, and the outer conductor is soldered to the other.
A male SMA connector is soldered to the free end of the coax cable so it can feed into an RF dongle.
The entire design and construction process can be completed in under an hour.
\Cref{fig:antennas} shows completed antennas.

The use of custom antennas may seem counter-intuitive given our goal of making a hassle-free system.
However, there are several reasons for our decision to use custom antennas.
First, researchers might be interested in a frequency that is not commonly used and for which no commercial antennas exist.
Second, custom antennas can be significantly less expensive than their COTS counterparts.
Our system uses two antennas at once, so a researcher interested in three separate frequencies would have to purchase six antennas.
Antennas range from tens to hundreds of dollars. In contrast, the material cost of our antennas---copper wire, styrofoam, coaxial cable---is negligible.
Finally, most COTS antennas are not designed specifically for UAVs and can be heavy.
For example, the commercially available \SI{217}{\mega\Hz} antenna in \Cref{fig:antennas} has a mass of \SI{0.54}{\kilogram}, whereas our Moxon is only \SI{0.12}{\kilogram}.

\subsection{RTL-SDR Dongles}
Although it makes sense to use custom antennas, RF electronics are more complicated and there is less incentive to deviate from COTS components.
There exist good COTS components that cover most of the VHF and UHF spectrums.

\begin{figure}
	\centering
	\includegraphics[width=3.2in]{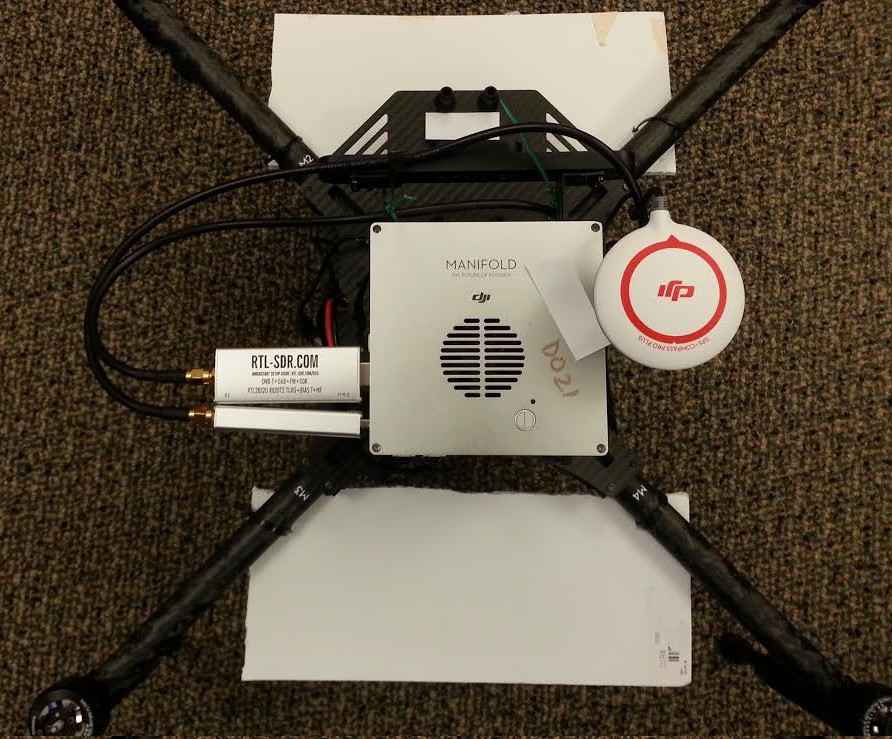}
	\caption{The Manifold onboard computer (center) has two RTL-SDR dongles in its USB ports (left). Each dongle is plugged into an antenna. The antennas (\SI{432.7}{\mega\Hz} in this picture) lie against the underside of styrofoam board.}
	\label{fig:manifold}
\end{figure}

We use RTL-SDR dongles for our RF analysis.
RTL-SDR refers to any software-defined radio (SDR) based on the RTL2832U chipset.
These chipsets were originally designed to receive television broadcasts, but they can be modified into SDRs if combined with a tuner.
We use dongles with the Rafael Micro R820T tuner, which are lightweight and low-cost (\SI{30}{\gram}, \$20 USD).
The dongles have a female SMA connector on one end and a USB connector on the other.
\Cref{fig:manifold} shows two dongles plugged into our UAV's onboard computer.

These RTL-SDR dongles are versatile RF analyzers.
With a stable sample rate of \SI{2.4}{\mega\Hz}, they can analyze \SI{2.4}{\mega\Hz} of bandwidth at once.
They have a frequency range of \num{24}-\SI{1766}{\mega\Hz}, meaning they can cover all of the VHF and most of the UHF frequency ranges.
There exist open-source C and Python libraries to tune the dongle and read samples.
To measure signal power, the dongle is tuned to a frequency of interest.
The dongle then reads RF samples, which can be converted into a periodogram, or an estimate of the power spectral density.
Our signal strength estimate is simply the largest density in the sampled bandwidth.

\begin{figure}
	\centering
	\includegraphics[width=3.3in]{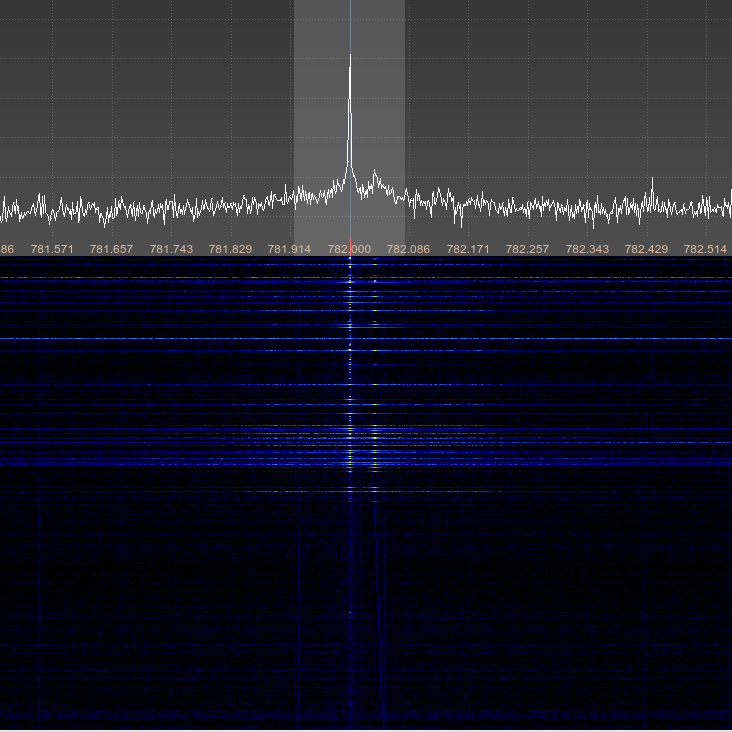}
	\caption{Using RTL-SDR dongle with open-source gqrx radio software to analyze emissions from cell phone placing voice call over LTE connection at \SI{782}{\mega\Hz}.
		The lower half of the waterfall plot corresponds to time before the call is placed; once the call is placed, emissions are logged.
	}
	\label{fig:analyzer}
\end{figure}

Because each dongle can analyze \SI{2.4}{\mega\Hz} at once, it can serve as a makeshift spectrum analyzer~\cite{nika2014,zhang2015b}. 
Thus, our measurement device doubles as a troubleshooting or exploratory device.
For example, cell phones radiate at many frequencies and it is not always clear which frequency is used in which region.
The RTL-SDR dongle can be used to check the spectrum for emissions, a technique we leveraged to determine the frequency at which our cell phone operates.
As shown in \Cref{fig:analyzer}, we tuned the dongle to the \SI{700}{\mega\Hz} region and found emissions, suggesting our cell phone operates in this band.

The simplicity and versatility of RTL-SDR dongles makes them an attractive tool to measure signal strength.
However, these dongles are not without drawbacks.
One drawback is that the R820T tuner has a maximum frequency of \SI{1766}{\mega\Hz}.
This upper limit covers most RF sources of interest, including wildlife radio-tags, the GPS frequency, ADS-B, and most cellular bands.
A comparably priced dongle (\$30 USD) with the Elonics E4000 tuner can reach up to \SI{2100}{\mega\Hz}, covering all cellular bands.
However, some commonly used frequencies, like WiFi (\SI{2.4}{\giga\Hz}), are outside this range.
For these cases, one can resort to SDRs like the HackRF One, which can reach up to \SI{6}{\giga\Hz}.
Unfortunately, this SDR costs \$300 USD, about an order of magnitude more than an RTL-SDR dongle.

\subsection{Manifold Computer}
The RTL-SDR dongles plug into the USB ports of a Manifold computer carried by the UAV.
This onboard computer can be seen in \Cref{fig:manifold}.
The Manifold is manufactured by DJI and retails for about \$500 USD.
The Manifold has \num{2} GB RAM and four ARM Cortex-A15 cores that clock up to \SI{2.3}{\giga\Hz}.
The Manifold is designed to parse video in flight, but our application does not use video so a cheaper alternative could probably be used.
For example, ODROID computers retail for under \$100 USD and have been used in previous localization tasks\cite{Perkins2015,bayram2016}.

The Manifold runs Ubuntu and ROS\cite{quigley2009}.
In flight, the Manifold estimates signal strength from the RTL-SDR dongles and queries the M-100's flight controller over serial.
The Manifold filters the strength measurements and UAV position to estimate the RF source's location.
The Manifold then computes and provides velocity commands to the flight controller.

\subsection{Transmitters}
We use three RF sources in our experiments.
The first is a Baofeng UV-5R radio.
This radio is popular with amateur radio enthusiasts and can transmit and receive in portions of the VHF and UHF bands.
For our tests, we set the radio to \SI{432.7}{\mega\Hz}, which is in the middle of the \SI{70}{\centi\meter} amateur band (\num{420}-\SI{450}{\mega\Hz}).
The radio is set to its low power setting (\SI{1}{\watt}) and we radiate constantly for the duration of a flight.

The second transmitter is a Nitehunters RATS-8 tracking collar.
The collar pulses once a second at \SI{217.335}{\mega\Hz} and is designed to be worn by hunting dogs so their owners can find them.
Although this collar is not a wildlife radio-tag, we will refer to it as such because it operates similarly and serves as a substitute.
The \num{216}-\SI{220}{\mega\Hz} band is commonly used for wildlife tracking, and wildlife transmitters typically pulse as well.
However, this collar retails for \$90 USD, which is a discount compared to collars sold for wildlife.
Wildlife collars are ruggedized and often sell for hundreds of dollars.

\begin{figure}
	\centering
	\includegraphics[width=3in]{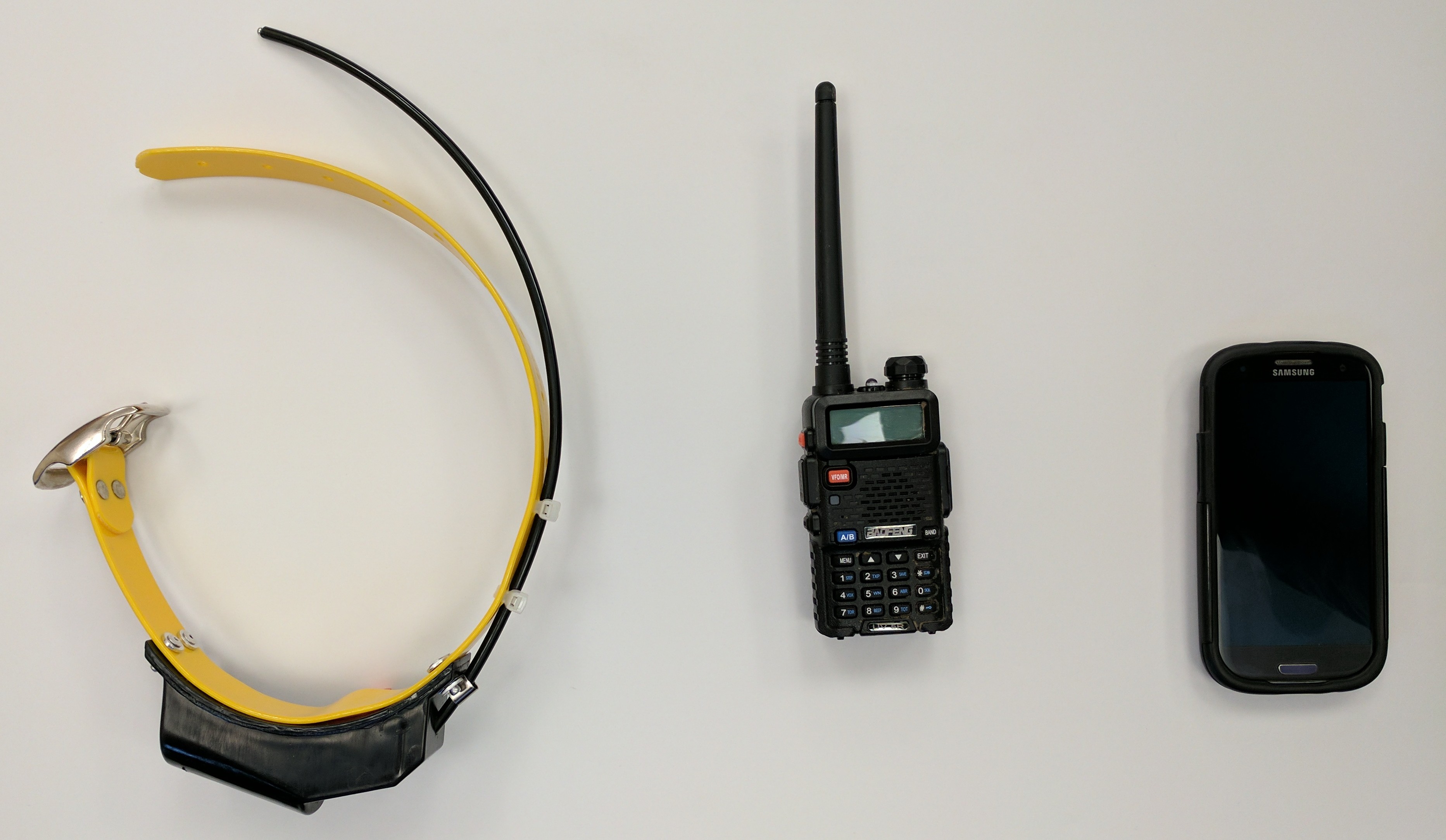}
	\caption{Transmitters used in experiments. From left to right: wildlife collar, Baofeng UV-5R, Samsung Galaxy S3.}
	\label{fig:transmitters}
\end{figure}

The third transmitter is a Samsung Galaxy S3 cell phone.
To have it transmit, we place a voice call over Verizon's LTE network.
In our region, this network operates in the \SI{700}{\mega\Hz} band.
Using an RTL-SDR dongle, we found the phone uplink frequency is \SI{782}{\mega\Hz}.
\Cref{fig:transmitters} shows the three transmitters.

\section{Sensing Modality}
\label{sec:modality}
In the previous section, we show how each antenna connects to an RTL-SDR dongle that estimates the signal strength received by each antenna.
This section discusses how these measurements are woven into a sensing modality with a corresponding sensing model.

\begin{figure}
	\centering
	\begin{tikzpicture}[]
	\begin{polaraxis}[width=2.7in]\addplot+ [mark = {none}]coordinates {
(0.0, 14.7976)
(0.8838446947688077, 15.1452)
(6.4882886636204695, 15.142)
(11.404151954283904, 15.2181)
(17.602638565127204, 14.8651)
(23.943848962737594, 14.6113)
(29.647357334367374, 14.303)
(35.11526546063935, 14.3041)
(40.93405294064945, 12.8296)
(46.9237537309471, 12.6318)
(52.116470228217736, 11.182)
(57.53607801236019, 10.675799999999999)
(62.71217873357206, 8.1401)
(67.78801184063602, 7.5411)
(74.24753802294092, 6.799300000000001)
(80.68872955580163, 3.0129)
(86.13125565174931, 0.1858000000000004)
(91.6906651379037, 1.7018)
(97.73708874991927, 0.35860000000000003)
(103.65860756259634, 0.19040000000000035)
(110.12672811182819, 0.0)
(116.09637537929625, 1.46)
(122.68195227652994, 4.552)
(129.0412108446869, 5.2575)
(135.46865139046452, 5.6513)
(141.61305078544746, 6.9752)
(148.51203559661766, 6.8634)
(155.03000347402593, 6.2161)
(161.0981994822565, 5.0286)
(166.24366456152268, 5.9893)
(171.73890277462243, 5.955)
(177.6535460937579, 6.1312999999999995)
(183.48682940598482, 5.0127)
(189.95609587080693, 3.561)
(195.67192283503203, 4.4219)
(201.7888202558487, 4.0646)
(208.65228168372082, 5.7673000000000005)
(213.69774802764283, 6.2104)
(219.26632483851935, 6.3408)
(225.71496482271675, 7.6649)
(231.11280021064422, 7.0021)
(237.3912717296878, 7.8479)
(241.77669069361912, 6.8073)
(247.360164407169, 5.5224)
(253.47820774357592, 6.4246)
(259.68219474925246, 4.7012)
(267.3999362496647, 2.3812)
(272.81897107601196, 2.3654)
(280.2914866401082, 3.1166)
(285.9185624418675, 4.4514000000000005)
(291.28408301014963, 5.8733)
(297.75988119383624, 8.2553)
(304.65823575143185, 9.9982)
(309.5145114314017, 11.01)
(313.86790205458516, 11.6226)
(318.9093576939413, 12.3339)
(325.51791019875975, 13.0228)
(332.52203276533646, 13.8011)
(337.5066796393054, 14.3124)
(342.4469721314798, 14.6362)
(349.5437494219805, 14.9136)
(356.9347330918296, 15.362400000000001)
(0.0, 14.7976)
};
\addlegendentry{Front Antenna}
\addplot+ [mark = {none}]coordinates {
(0.0, 4.4818)
(0.8838446947688077, 4.8138000000000005)
(6.4882886636204695, 4.4508)
(11.404151954283904, 4.4536)
(17.602638565127204, 4.2149)
(23.943848962737594, 2.2395)
(29.647357334367374, 1.6773000000000002)
(35.11526546063935, 0.9241999999999999)
(40.93405294064945, 1.8078000000000003)
(46.9237537309471, 3.5092)
(52.116470228217736, 4.5425)
(57.53607801236019, 4.9929)
(62.71217873357206, 5.4739)
(67.78801184063602, 6.3684)
(74.24753802294092, 5.9663)
(80.68872955580163, 5.0125)
(86.13125565174931, 5.1396)
(91.6906651379037, 5.6837)
(97.73708874991927, 4.7754)
(103.65860756259634, 5.9526)
(110.12672811182819, 7.7706)
(116.09637537929625, 8.929400000000001)
(122.68195227652994, 10.7532)
(129.0412108446869, 11.8154)
(135.46865139046452, 12.4748)
(141.61305078544746, 13.5923)
(148.51203559661766, 13.9358)
(155.03000347402593, 14.2985)
(161.0981994822565, 14.6456)
(166.24366456152268, 15.0308)
(171.73890277462243, 15.2165)
(177.6535460937579, 15.2058)
(183.48682940598482, 14.9659)
(189.95609587080693, 14.7166)
(195.67192283503203, 14.0853)
(201.7888202558487, 13.7256)
(208.65228168372082, 13.0231)
(213.69774802764283, 11.9382)
(219.26632483851935, 11.2146)
(225.71496482271675, 10.6369)
(231.11280021064422, 8.7154)
(237.3912717296878, 7.6878)
(241.77669069361912, 6.927)
(247.360164407169, 5.7794)
(253.47820774357592, 4.4872)
(259.68219474925246, 0.34309999999999974)
(267.3999362496647, 0.9800000000000004)
(272.81897107601196, 1.7021000000000002)
(280.2914866401082, 2.6226000000000003)
(285.9185624418675, 5.24)
(291.28408301014963, 6.3428)
(297.75988119383624, 7.3937)
(304.65823575143185, 8.0627)
(309.5145114314017, 8.0143)
(313.86790205458516, 7.8725000000000005)
(318.9093576939413, 8.5583)
(325.51791019875975, 8.732099999999999)
(332.52203276533646, 7.756)
(337.5066796393054, 6.2595)
(342.4469721314798, 5.4311)
(349.5437494219805, 4.7447)
(356.9347330918296, 4.6111)
(0.0, 4.4818)
};
\addlegendentry{Rear Antenna}
\end{polaraxis}

\end{tikzpicture}
	\caption{Signal strengths as functions of relative bearing to RF source. The front antenna receives higher strength when the UAV faces the RF source. The transmitter used was the UV-5R radio.}
	\label{fig:two_432}
\end{figure}
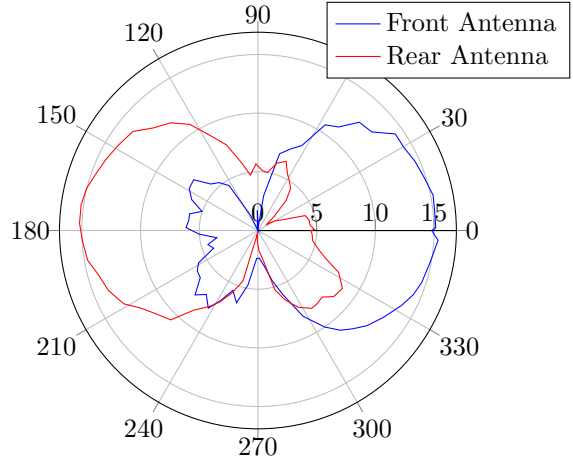

We mount one antenna facing forward and another facing back.
Strength measurements are made simultaneously.
Because the antennas are directional, the antenna facing the RF source should receive higher strength.
If higher strength is measured from the front-facing antenna, the RF source lies somewhere in front of the UAV; higher strength from the rear-facing antenna implies the source is behind the UAV.
\Cref{fig:two_432} shows strength measurements made while our UAV rotated in place.
The resulting pattern validates the idea of measuring higher strength from the antenna facing the RF source.

We call the resulting sensing modality field-of-view (FOV) sensing because it is conceptually similar to detection-based modalities using cameras and laser scanning\cite{wehr1999,Symington2010}.
The output is 1 if the front antenna measurement is higher and 0 if not.
This model condenses two continuous, real-valued strength measurements into an observation $z\in\{0,1\}$.
The reduction is useful if planning requires an expectation over observations, which is often the case in information-theoretic control.
We denote the set of possible observations as $\mathcal{Z}$; in this sensing modality, $\mathcal{Z} = \{0,1\}$.

A sensor model is a probabilistic model that defines the probability---at time $t$---of making measurement $\zt$ if the UAV state is $\xt$ and the RF source location is $\theta_t$.
We model the UAV state as a point in the special Euclidean group $\se$, meaning it consists of a 2D position and a UAV heading.
The RF source location $\theta_t\in\mathbb{R}^2$ only consists of a 2D position and is assumed constant, so we write $\theta_t=\theta$.
Explicitly, $\xt$ and $\theta$ are
\begin{equation}
	\begin{aligned}
		\xt &= \left[\xtn, \xte, h_t\right]^\top\text,\\
		\theta &= \left[\thetan, \thetae\right]^\top\text,
	\end{aligned}
\end{equation}
where $\xtn$ and $\xte$ represent the north and east components of the UAV position, respectively.
Likewise, $\thetan$ and $\thetae$ represent the north and east components of the RF source position.
The UAV heading, denoted $h_t$, is measured east of north and defines the direction the front-facing antenna is pointing.

We define the bearing $\beta_t$ as the angle, measured east of north, of a ray pointing from the UAV position to the transmitter position:
\begin{equation}
	\beta_t = \arctan\left(\frac{\thetae - \xte}{\thetan - \xtn}\right)\text.
\end{equation}
We define the quantity $\beta_t - h_t$ as the relative bearing. 
When the relative bearing is \ang{0}, the front antenna points directly at the RF source, and a measurement of 1 is expected.
When the relative bearing is \ang{180}, the rear antenna points directly at the RF source, and a measurement of 0 is expected.
We would expect a measurement of 1 if the relative bearing is in the interval $[\ang{-90}, \ang{90}]$, meaning the front antenna is pointed more closely to the RF source than the rear antenna.
However, the front and rear antenna gains become similar at relative bearings near $\pm\ang{90}$, so mistakes are more likely.
Therefore we define a cone width $\alpha \leq \ang{180}$ over which we are confident the proper measurement will be returned.
Intuitively, this setup can be thought of as two cones of width $\alpha$ and with vertices at the UAV position; one cone is centered along the UAV heading and the other in the opposite direction.
If the RF source lies in the front cone, we expect a measurement of 1; if the RF source lies in the rear cone, we expect a measurement of 0.
If the RF source lies between these cones, we expect either measurement to be equally likely.
We also define a mistake rate $\mu$, which is the probability the UAV misidentifies the cone containing the RF source, if the source lies in one.
We set $\mu=0.1$ in flight tests and simulations.
Mathematically, we have

\begin{equation}
	P(z_t = 1\mid x_t, \theta) = \begin{cases}
		1-\mu, & \text{if }\beta_t - h_t \in \left[-\frac{\alpha}{2},\ \frac{\alpha}{2}\right]\\
		\hfil \mu, & \text{if }\beta_t - h_t \in \left[\ang{180}-\frac{\alpha}{2},\ \ang{180}+\frac{\alpha}{2}\right]\\
		\hfil 0.5, & \text{otherwise}
	\end{cases}
	\label{eq:sensor}
\end{equation}

\Cref{fig:two_432} indicates that $\alpha=\ang{120}$ is an appropriate cone width.
If the RF source lies between front and rear cones of with \ang{120}, we deem either measurement equally likely.
A benefit of limiting cone width is that the resulting uncertainty region encodes uncertainty caused by imperfect antennas.
Ideally, the front and rear cones would have a width $\alpha=\ang{180}$, but this would require perfectly constructed and placed antennas.
In reality, some antennas might have larger side and back lobes in their gain patterns; or, the front-facing antenna might not be placed exactly along the UAV's heading axis.
By including the uncertainty region, we alleviate the pressure on operators to construct and align their antennas perfectly.

\begin{figure}
	\centering
    \begin{tikzpicture}[]
	\begin{polaraxis}[width=2.7in]\addplot+ [mark = {none}]coordinates {
(0.0, 24.465500000000002)
(359.96963323685804, 24.1641)
(1.906058697061713, 24.3288)
(8.20005737235332, 24.2789)
(14.650588117274665, 24.69)
(20.500200726662804, 24.557000000000002)
(26.65720519313863, 19.237900000000003)
(32.81472532163007, 14.776600000000002)
(39.04850613265343, 17.7253)
(45.23530440447607, 20.6367)
(51.682798473083224, 21.761000000000003)
(58.07586155115294, 19.651600000000002)
(63.79054259978777, 17.8074)
(70.60988627743482, 16.070400000000003)
(76.369830991885, 7.794100000000002)
(81.84157793538435, 3.686300000000001)
(87.49437954214507, 8.272100000000002)
(93.10363635647583, 10.244700000000002)
(99.70697494535855, 9.244600000000002)
(105.18215963562872, 10.110700000000001)
(111.06127957146609, 9.071600000000002)
(116.98509021532368, 4.572900000000002)
(123.19251496777102, 0.6085000000000029)
(128.9289684126208, 0.0)
(135.18624049324453, 6.793500000000002)
(141.96433120964215, 13.379900000000001)
(147.45326688699544, 15.536700000000002)
(152.95339284144765, 15.0042)
(159.06685251549354, 15.626400000000002)
(164.85601807749535, 14.681700000000001)
(170.6417458927264, 7.795700000000002)
(176.07223987497633, 6.205000000000002)
(182.2481519486915, 12.0928)
(187.66718677503883, 12.981500000000002)
(192.910323558281, 13.3793)
(199.27244691541364, 13.280000000000001)
(205.1956846014761, 12.807800000000002)
(211.7388626218701, 8.485500000000002)
(217.966340897147, 12.701)
(223.67357349444512, 16.636300000000002)
(230.17607151138486, 16.195200000000003)
(236.9192118022795, 16.579400000000003)
(242.6533734159488, 16.9116)
(248.4568629228289, 15.646)
(254.76283641603874, 12.185300000000002)
(260.531948455211, 9.111700000000003)
(266.1664154125275, 11.6903)
(272.82246611856226, 12.0347)
(279.8084259230334, 8.951800000000002)
(285.215543227242, 12.508500000000002)
(290.839009394892, 11.274400000000002)
(297.14286294425983, 7.874800000000002)
(302.7659853372328, 13.364100000000002)
(308.82151897819097, 17.956000000000003)
(314.29527127397324, 20.5589)
(321.30111271393537, 20.1183)
(327.4868338251032, 22.382600000000004)
(334.4095791960375, 21.9645)
(340.47978628612896, 13.8808)
(346.29754244210784, 13.369200000000001)
(351.59711556817035, 21.6387)
(356.79974423529677, 24.235500000000002)
(0.0, 24.465500000000002)
};
\addlegendentry{Front Antenna}
\addplot+ [mark = {none}]coordinates {
(0.0, 13.919100000000002)
(359.96963323685804, 14.451200000000002)
(1.906058697061713, 13.053500000000001)
(8.20005737235332, 14.352400000000001)
(14.650588117274665, 12.339200000000002)
(20.500200726662804, 11.140400000000001)
(26.65720519313863, 6.802400000000002)
(32.81472532163007, 2.176400000000001)
(39.04850613265343, 6.674300000000002)
(45.23530440447607, 10.705300000000001)
(51.682798473083224, 12.855200000000002)
(58.07586155115294, 14.1582)
(63.79054259978777, 14.590300000000003)
(70.60988627743482, 13.512200000000002)
(76.369830991885, 5.841300000000002)
(81.84157793538435, 3.790000000000001)
(87.49437954214507, 9.031900000000002)
(93.10363635647583, 13.610700000000001)
(99.70697494535855, 14.580400000000001)
(105.18215963562872, 15.232200000000002)
(111.06127957146609, 16.340500000000002)
(116.98509021532368, 13.068400000000002)
(123.19251496777102, 9.486900000000002)
(128.9289684126208, 8.199500000000002)
(135.18624049324453, 15.822700000000001)
(141.96433120964215, 21.606700000000004)
(147.45326688699544, 23.232000000000003)
(152.95339284144765, 23.6916)
(159.06685251549354, 24.0714)
(164.85601807749535, 23.1365)
(170.6417458927264, 19.230800000000002)
(176.07223987497633, 17.4203)
(182.2481519486915, 22.8424)
(187.66718677503883, 24.1967)
(192.910323558281, 23.2465)
(199.27244691541364, 23.9371)
(205.1956846014761, 21.666200000000003)
(211.7388626218701, 17.109900000000003)
(217.966340897147, 18.7113)
(223.67357349444512, 20.5208)
(230.17607151138486, 18.040300000000002)
(236.9192118022795, 17.8641)
(242.6533734159488, 16.8667)
(248.4568629228289, 15.176800000000002)
(254.76283641603874, 11.662500000000001)
(260.531948455211, 7.745000000000001)
(266.1664154125275, 10.1462)
(272.82246611856226, 10.314300000000003)
(279.8084259230334, 10.670100000000001)
(285.215543227242, 13.168300000000002)
(290.839009394892, 10.720500000000001)
(297.14286294425983, 7.3808000000000025)
(302.7659853372328, 12.759300000000001)
(308.82151897819097, 16.7586)
(314.29527127397324, 18.2891)
(321.30111271393537, 16.093700000000002)
(327.4868338251032, 16.7356)
(334.4095791960375, 15.686600000000002)
(340.47978628612896, 7.226100000000002)
(346.29754244210784, 5.5576000000000025)
(351.59711556817035, 12.132100000000001)
(356.79974423529677, 13.462900000000001)
(0.0, 13.919100000000002)
};
\addlegendentry{Rear Antenna}
\end{polaraxis}

\end{tikzpicture}
	\caption{Strength measurements while rotating UV-5R so received strength changes. Both front and rear measurements are affected equally.}
	\label{fig:robust1}
\end{figure}
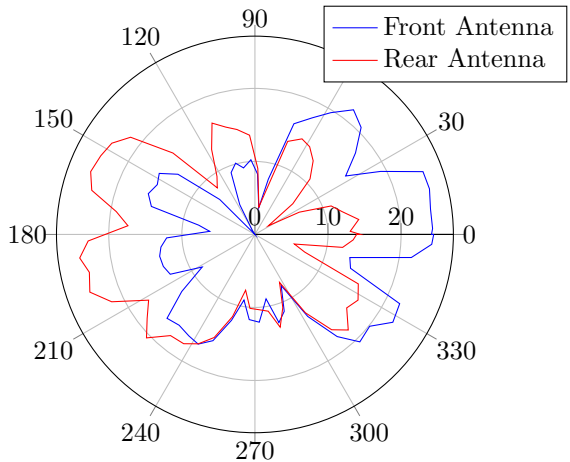

Not only is FOV sensing robust to antenna construction and placement, but it is also robust to time-varying factors like those discussed in \Cref{sec:background}.
For example, transmitter strength and orientation can change during a rotation.
We tested this by manually rotating our UV-5R radio as the UAV rotated in place and made strength measurements.
The radio's antenna is a dipole, so it has low gain along its antenna's axis.
The resulting strength measurements can be seen in \Cref{fig:robust1}.
The patterns are distorted because changes in the radio's orientation affect the strength reaching the UAV's position.
Traditional rotate-for-bearing approaches would have difficulty estimating bearing from these patterns.
However, our two-antenna approach is not affected because both antennas are affected equally.
The front-facing antenna measures greater strength in the front cone, and the rear antenna measures greater strength in the rear cone.

\section{Localization Algorithm}
\label{sec:algorithm}
We implement a greedy, information-theoretic controller to provide velocity commands to the UAV.

\subsection{Dynamic Model}
We apply assumptions and restrictions to the UAV dynamic model in order to simplify implementation on a real UAV.
First, we restrict the UAV's motion to a horizontal plane, constraining it to a constant altitude.
Moxon antennas have roughly constant gain over elevation angle to the transmitter, so changes in UAV altitude would not yield much information about the transmitter location.
It is possible to use antennas that are sensitive to elevation angle, but this scheme would require enhanced antenna modeling.
Such a scheme would also be vulnerable to uncertainty in the transmitter's altitude.
Because our goal is a simple, robust system, we use the Moxon antennas, leading to the reasonable constant-altitude restriction.
Further, this restriction is common in previous UAV localization work\cite{cliff2015,Perkins2015,Perkins2016,bayram2016}.

We also assume the dynamics are deterministic and that the current UAV state is known.
The UAV's magnetometer provides heading information, and GPS is used for 2D position.
We neglect any GPS position error.
The assumption of known UAV position might seem questionable in certain applications, like GPS jamming.
However, there are many alternative methods for a UAV to estimate its own position, from vision to satellite navigation systems at other frequencies\cite{Perkins2016}.
Because we assume no noise in the dynamics, knowledge of the current state is enough to determine the UAV's position at future time steps.

We model the UAV's planar motion as a single integrator for two reasons.
First, the UAV's flight controller accepts velocity commands; second, the UAV is maneuverable and can change velocities quickly.
Of course, the UAV will spend some time accelerating to commanded velocities, but the approximation is not unreasonable.
Further, the UAV does not plan a long trajectory where propagation errors accumulate---it makes single-step plans and errors in the dynamics model will not be critical.

Measurements are received at discrete time steps spaced $\Delta t$ apart.
Using the simple forward Euler technique, we can propagate the UAV's state at time $t$ to the next time step $t+\Delta t$:
\begin{equation}
	\xtt = \xt + \ut \Delta t\text.
\end{equation}

\subsection{Belief and Filtering}

The search area is modeled as a square, with the UAV starting in the center.
We split the search area into a grid, where the density of each grid cell represents the probability that the RF source is in that cell.
This technique is called discrete or histogram filtering\cite{Thrun2005}.
It is common in UAV localization\cite{cliff2015,Perkins2015,Perkins2016} because it can handle non-Gaussian priors and non-linear dynamic and measurement models.
We refer to this distribution over RF source locations as a belief.
The belief is initialized to a uniform distribution with each cell weighted equally, indicating the RF source is equally likely to be in any cell.

The belief $\bt$ is computed from the preceding belief $b_{t-\dt}$, the UAV state $x_t$, the observation $z_t$, the measurement model from \cref{eq:sensor}, and Bayes' rule:
\begin{equation}
\bt(\theta_i) \propto b_{t-\dt}(\theta_i)P(\zt\mid \xt, \theta_i),
\label{eq:belief_update}
\end{equation}
where $\theta_i$ is a cell and $\bt(\theta_i)$ represents the probability the RF source is in cell $\theta_i$.
For simplicity, the center of a cell is used in the measurement model.

Each flight test used a stationary transmitter, so our model assumes a stationary transmitter.
However, we could extend to moving transmitter applications by including a stochastic motion model in \cref{eq:belief_update}.

\subsection{Algorithm}
At time $t$, the UAV makes measurement $\zt$, yielding the current belief $\bt$.
This belief is used to generate a velocity command with the goal of minimizing belief uncertainty as quickly as possible.
One measure of uncertainty is entropy, which captures the spread of a probability distribution.
The entropy of discrete distribution $\bt$, denoted $H(\bt)$, is computed with

\begin{equation}
	H(\bt) = -\sum_{\theta_i\in\Theta} \bt(\theta_i)\log{\bt(\theta_i)}\text,
\end{equation}
where by convention $0\log{0} = 0$.
Entropy is minimized when the probability is concentrated in a single cell and maximized when all cells have equal probability.

A common approach to automated localization is greedily minimizing entropy\cite{Hoffmann2006,cliff2015}.
At each step, the UAV picks the action that minimizes the expected entropy of the belief after receiving the next measurement.
This method is called greedy or myopic because it only plans over a single step.
Planning myopically is generally suboptimal, as it can lead to short-term gain at the expense of long-term performance.
However, multi-step planning for information gathering is computationally challenging\cite{Dressel2017}.
Further, greedy entropy minimization has performed well in a variety of tasks\cite{Hoffmann2006,cliff2015,Dressel2017}.

The set of possible control actions is technically infinite, but we discretize it to make the optimization easier.
At each time step, the UAV reasons over this discrete set, evaluating the expected reduction in belief entropy after taking a specific action and making a new measurement.
The action set is the Cartesian product of the velocity action set and the heading command set.
The velocity action set consists of eight actions: move \SI{5}{\meter\per\second} in one of eight directions spaced \ang{45} apart (north, northeast, etc.).
The heading action set consists of three actions: rotate \SI{10}{\degree\per\second} in either direction or do not rotate.
The velocity set only includes speeds of $\SI{5}{\meter\per\second}$, because this speed was empirically found to move the UAV quickly without maneuvering violently.
This restriction could be lifted in the future.

To evaluate an action $\ut$, the UAV considers the resulting state $\xtt$, which is fixed by knowledge of $\xt$ and the deterministic dynamic model.
The measurement received at this new state, $\ztt$, will lead to a new belief $\btt$.
In myopic entropy reduction, the objective is to minimize $\exz H(\btt)$, the expected entropy after taking measurement $\ztt$ at the next step.
The objective is equivalent to $H(\bt \mid \ztt)$, the conditional entropy between distribution $\bt$ and $\ztt$.
This conditional entropy expresses what the uncertainty in the transmitter location would be if $\ztt$ were known.
We treat $\ztt$ as a random variable because it is an unknown future quantity---as opposed to $\xtt$, which is specified by $\ut$.\footnote{We abuse notation and use $\bt$ as an argument to information-theoretic quantities, even though it is a distribution and not a random variable. When we do so, we imply a random variable describing the transmitter location and having distribution $\bt$.}
Conditional entropy can be expanded:\cite{thomas2006} 
\begin{equation}
H(\bt \mid \ztt) = H(\bt) - I(\ztt; \bt)\text,
\end{equation}
where $I(\ztt; \bt)$ is the mutual information between the target and sensor distributions.
The entropy of the current belief, $H(\bt)$, cannot be changed, so maximizing the mutual information $I(\ztt; \bt)$ minimizes the posterior entropy.
This result satisfies intuition, as the mutual information between $\ztt$ and $\bt$ expresses the reduction in uncertainty of belief $\bt$ if we knew $\ztt$~\cite{thomas2006}.
Mutual information is symmetric so $I(\ztt;\bt) = I(\bt; \ztt)$.
Re-expanding leads to
\begin{equation}
I(\bt; \ztt) = H(\ztt) - H(\ztt \mid \bt)\text.
\label{eq:objective}
\end{equation}
The UAV evaluates \cref{eq:objective} for each possible action $\ut$ and corresponding future state $\xtt$, selecting the maximizing action.

We can break down the terms in \cref{eq:objective} to gain an intuitive understanding of greedy entropy minimization~\cite{Hoffmann2006}.
The term $H(\ztt)$ represents uncertainty in $\ztt$, the measurement to be received at the next state $\xtt$.
We want this term to be large; intuitively, we learn when we sample from outcomes we are unsure of.
The term $H(\ztt \mid \bt)$ represents the uncertainty $\ztt$ would have if the transmitter's location were known.
The UAV is evaluating an action $\ut$ so $\xtt$ is known. If the transmitter's location is also known, any uncertainty in the measurement to be received is due to sensor noise.
We might learn when we sample from outcomes we are unsure of, but not if the outcomes are very noisy.
Therefore, picking $\xtt$ to maximize $H(\ztt) - H(\ztt \mid \bt)$ is equivalent to picking $\xtt$ such that the UAV is uncertain about which measurement it will receive, but not simply because of sensor noise.

We now show how to compute the objective function in \cref{eq:objective} using the current belief $\bt$, knowledge of $\xtt$, and the measurement model.
Consider the first term $H(\ztt)$:
\begin{equation}
	H(\ztt) = -\sum_{z\in\mathcal{Z}} P(\ztt=z )\log P(\ztt=z),
\label{eq:firstterm}
\end{equation}
Because $\xtt$ is implied by $\ut$, we can write $P(\ztt\seq z) = P(\ztt\seq z \mid \xtt)$.
The measurement model (\cref{eq:sensor}) depends on transmitter location, so we apply the laws of total and conditional probability:
\begin{equation}
P(\ztt \seq z \mid \xtt) = \sum_{\theta_i\in\Theta} P(\ztt \seq z \mid \xtt, \theta_i) \bt(\theta_i).
\end{equation}
The second term in the objective from \cref{eq:objective} is $H(\ztt \mid \bt)$:
\begin{equation}
	H(\ztt \mid \bt) = -\sum_{\theta_i\in\Theta} \bt(\theta_i)\sum_{z\in\mathcal{Z}} P(\ztt \seq z\mid \xtt,\theta_i)\log P(\ztt \seq z\mid \xtt, \theta_i).
\label{eq:secondterm}
\end{equation}

\section{Simulations}
\label{sec:simulations}

To evaluate the performance of FOV sensing, we ran simulations with a single UAV searching for a single, stationary transmitter in a $\SI{200}{\meter} \times \SI{200}{\meter}$ search area that is discretized into 1600 $\SI{5}{\meter} \times \SI{5}{\meter}$ cells.
In each simulation, the transmitter is initialized to a random location in the search area and the UAV starts in the center of the search area.
For evaluation, we measure time to localization, which for these simulations we define as concentrating 50\% of the belief in a single cell.
This concentration is equivalent to achieving a belief max-norm of 0.5.
Denoted $\norm{b}_\infty$, the max-norm indicates the highest probability mass in any cell in belief $b$.
When the max-norm is 1, the belief is concentrated in one cell and we are certain---according to our model---the transmitter is in that cell. 
A max-norm of 0.5 or greater means most of the belief is in one cell, making it a reasonable criteria for localization.

Our localization algorithm minimizes belief entropy at the next time step, but we use the belief max-norm for evaluation.
The reason for this mismatch is that optimizing entropy is common in robotic localization, but the max-norm is much easier to understand. A max-norm of 0.5 is more intuitive than a belief entropy of 4.2~nats.
Further, these metrics are related and reach their extrema when the belief is uniform or concentrated in a single state, so it is not unreasonable to optimize for one and evaluate with the other.

%

\subsection{Comparison to Other Modalities}

We compare FOV sensing to two other sensing modalities.
The first is an instantaneous bearing (IB) modality that provides bearing estimates in real-time. 
This modality might be implemented by a complex method like beam-steering.
We include the IB modality to estimate the decrease in performance caused by avoiding the complexities of beam-steering.
The second modality is the rotate-for-bearing (RFB) method in which the UAV rotates in place and estimates the bearing with a directional antenna.
The UAV then moves to the next measurement location and rotates again.

We assume the FOV sensor receives measurements sampled according to \cref{eq:sensor} at \SI{1}{\Hz}.
The cone width and mistake rate are assumed to be $\alpha = \ang{120}$ and $\mu = 0.1$.
The IB method also samples at \SI{1}{\Hz}. 
Bearing estimates for both the IB and RFB methods have zero-mean, Gaussian noise with a standard deviation of \ang{5}.
This noise level is roughly half that reported in some work~\cite{Perkins2015}.
When using the RFB method, the UAV takes \SI{24}{\second} to rotate.
With all modalities, the UAV moves at \SI{5}{\meter\per\second}.
We apply greedy, information-theoretic controllers for all modalities.
For the RFB modality, this greedy controller picks the measurement location that most reduces entropy.
The UAV then moves to this point and rotates to get a bearing estimate.
The controller then picks a new measurement location.
We also apply random controllers to the FOV and IB modalities to see the effect of the information-theoretic controllers.


Shown in \Cref{tab:simulations}, the simulation results show two important results.
First, FOV sensing can significantly outperform the RFB scheme.
Mean localization with FOV sensing occurs in roughly the same time a \textit{single} bearing measurement is made---or less, if you use \num{40} and \SI{45}{\second} per rotation as reported in some work~\cite{vonehr2016,cliff2015}.
FOV localization occurs in less than a third of the time that RFB takes. This estimate is conservative and the improvement would be even greater if longer rotation times or larger bearing standard deviations were used for the RFB method.
The performance of FOV is much closer to that of IB, despite being much less complex than beam-steering.

\begin{table}
	\caption{Mean time to concentrate 50\% of the belief in a single $\SI{5}{\meter}\times \SI{5}{\meter}$ cell in a $\SI{200}{\meter}\times \SI{200}{\meter}$ search area.}
	\centering
    \begin{tabular}{@{} lcc @{}}
    \toprule
	Sensor & Control & Localization Time (s) \\
    \midrule
	FOV & Greedy & 30.8 \\
	FOV & Random & 499 \\
    IB & Greedy & 17.5 \\
    IB & Random & 115 \\
	RFB & Greedy & 99.2 \\
    \bottomrule
    \end{tabular}
	\label{tab:simulations}
\end{table}

Second, greedy control improves FOV localization by an order of magnitude when compared to a random policy.
This stark improvement suggests each measurement is uninformative, so carefully selecting the control input in order to gather more information is needed for good performance.
In contrast, the performance of IB with random control is not nearly as poor; it performs roughly as well as the RFB method.
Each bearing estimate is informative so the effect of intelligent control is less noticeable.

\begin{figure}
	\centering
	\begin{tikzpicture}[]
\begin{axis}[legend pos = {south east}, ylabel = {Belief Max-norm}, xmin = {0}, xmax = {120}, xlabel = {Time, $s$}]\addplot+ [mark = {none}, thick, dashed]coordinates {
(0.0, 0.000625)
(1.0, 0.014447945986812835)
(2.0, 0.022330307041647134)
(3.0, 0.03230580147943695)
(4.0, 0.04461469680841376)
(5.0, 0.07891814788082117)
(6.0, 0.0927129906809922)
(7.0, 0.11727216205037828)
(8.0, 0.1143462022302574)
(9.0, 0.12626778635890382)
(10.0, 0.15959426817121142)
(11.0, 0.1386266700365974)
(12.0, 0.18813795835784983)
(13.0, 0.17005632084427622)
(14.0, 0.2889075404928286)
(15.0, 0.1986751897074832)
(16.0, 0.24559851931596688)
(17.0, 0.2296596248014155)
(18.0, 0.24964171340946945)
(19.0, 0.28205712124337423)
(20.0, 0.18359957983838432)
(21.0, 0.2131252336040479)
(22.0, 0.19617650716024002)
(23.0, 0.30213696906367266)
(24.0, 0.5143610555866618)
(25.0, 0.6177770936409674)
(26.0, 0.7369621076972279)
(27.0, 0.8227833421669412)
(28.0, 0.7492996240678821)
(29.0, 0.9703049834793057)
(30.0, 0.9687968423779789)
(31.0, 0.9995921819369438)
(32.0, 0.9995921819369438)
(33.0, 0.9995921819369438)
(34.0, 0.9995921819369438)
(35.0, 0.9995921819369438)
(36.0, 0.9995921819369438)
(37.0, 0.9995921819369438)
(38.0, 0.9995921819369438)
(39.0, 0.9995921819369438)
(40.0, 0.9995921819369438)
(41.0, 0.9995921819369438)
(42.0, 0.9995921819369438)
(43.0, 0.9995921819369438)
(44.0, 0.9995921819369438)
(45.0, 0.9995921819369438)
(46.0, 0.9995921819369438)
(47.0, 0.9995921819369438)
(48.0, 0.9995921819369438)
(49.0, 0.9995921819369438)
(50.0, 0.9995921819369438)
(51.0, 0.9995921819369438)
(52.0, 0.9995921819369438)
(53.0, 0.9995921819369438)
(54.0, 0.9995921819369438)
(55.0, 0.9995921819369438)
(56.0, 0.9995921819369438)
(57.0, 0.9995921819369438)
(58.0, 0.9995921819369438)
(59.0, 0.9995921819369438)
(60.0, 0.9995921819369438)
(61.0, 0.9995921819369438)
(62.0, 0.9995921819369438)
(63.0, 0.9995921819369438)
(64.0, 0.9995921819369438)
(65.0, 0.9995921819369438)
(66.0, 0.9995921819369438)
(67.0, 0.9995921819369438)
(68.0, 0.9995921819369438)
(69.0, 0.9995921819369438)
(70.0, 0.9995921819369438)
(71.0, 0.9995921819369438)
(72.0, 0.9995921819369438)
(73.0, 0.9995921819369438)
(74.0, 0.9995921819369438)
(75.0, 0.9995921819369438)
(76.0, 0.9995921819369438)
(77.0, 0.9995921819369438)
(78.0, 0.9995921819369438)
(79.0, 0.9995921819369438)
(80.0, 0.9995921819369438)
(81.0, 0.9995921819369438)
(82.0, 0.9995921819369438)
(83.0, 0.9995921819369438)
(84.0, 0.9995921819369438)
(85.0, 0.9995921819369438)
(86.0, 0.9995921819369438)
(87.0, 0.9995921819369438)
(88.0, 0.9995921819369438)
(89.0, 0.9995921819369438)
(90.0, 0.9995921819369438)
(91.0, 0.9995921819369438)
(92.0, 0.9995921819369438)
(93.0, 0.9995921819369438)
(94.0, 0.9995921819369438)
(95.0, 0.9995921819369438)
(96.0, 0.9995921819369438)
(97.0, 0.9995921819369438)
(98.0, 0.9995921819369438)
(99.0, 0.9995921819369438)
(100.0, 0.9995921819369438)
(101.0, 0.9995921819369438)
(102.0, 0.9995921819369438)
(103.0, 0.9995921819369438)
(104.0, 0.9995921819369438)
(105.0, 0.9995921819369438)
(106.0, 0.9995921819369438)
(107.0, 0.9995921819369438)
(108.0, 0.9995921819369438)
(109.0, 0.9995921819369438)
(110.0, 0.9995921819369438)
(111.0, 0.9995921819369438)
(112.0, 0.9995921819369438)
(113.0, 0.9995921819369438)
(114.0, 0.9995921819369438)
(115.0, 0.9995921819369438)
(116.0, 0.9995921819369438)
(117.0, 0.9995921819369438)
(118.0, 0.9995921819369438)
(119.0, 0.9995921819369438)
};
\addlegendentry{IB}
\addplot+ [mark = {none}, thick]coordinates {
(0.0, 0.000625)
(1.0, 0.0011249999999999854)
(2.0, 0.0014763779527558918)
(3.0, 0.0017962566897625576)
(4.0, 0.0021331812589573482)
(5.0, 0.008428563679911965)
(6.0, 0.004185020969882839)
(7.0, 0.0033789811350407053)
(8.0, 0.008928225395928285)
(9.0, 0.012932929022241215)
(10.0, 0.012200450717403711)
(11.0, 0.01633175742822627)
(12.0, 0.008694461433835085)
(13.0, 0.00916018376807362)
(14.0, 0.01302180026999201)
(15.0, 0.01732439835999159)
(16.0, 0.022880042668716426)
(17.0, 0.0341200417924645)
(18.0, 0.030003046598838908)
(19.0, 0.060061574594721306)
(20.0, 0.060266772273740846)
(21.0, 0.08710094197968478)
(22.0, 0.12437532160289554)
(23.0, 0.0960991538619177)
(24.0, 0.11650575921598247)
(25.0, 0.08287924972429868)
(26.0, 0.12534553382898125)
(27.0, 0.18149968170513417)
(28.0, 0.27608721291204735)
(29.0, 0.38584538630031573)
(30.0, 0.2126788183296446)
(31.0, 0.40753203232382973)
(32.0, 0.5271555306002489)
(33.0, 0.7444742444362137)
(34.0, 0.8202962481219663)
(35.0, 0.9198495055870181)
(36.0, 0.9542941211984349)
(37.0, 0.9719768448938283)
(38.0, 0.9878509197740023)
(39.0, 0.9931872985095692)
(40.0, 0.9981362903529591)
(41.0, 0.9988839938791495)
(42.0, 0.9995502677786328)
(43.0, 0.9995502677786328)
(44.0, 0.9995502677786328)
(45.0, 0.9995502677786328)
(46.0, 0.9995502677786328)
(47.0, 0.9995502677786328)
(48.0, 0.9995502677786328)
(49.0, 0.9995502677786328)
(50.0, 0.9995502677786328)
(51.0, 0.9995502677786328)
(52.0, 0.9995502677786328)
(53.0, 0.9995502677786328)
(54.0, 0.9995502677786328)
(55.0, 0.9995502677786328)
(56.0, 0.9995502677786328)
(57.0, 0.9995502677786328)
(58.0, 0.9995502677786328)
(59.0, 0.9995502677786328)
(60.0, 0.9995502677786328)
(61.0, 0.9995502677786328)
(62.0, 0.9995502677786328)
(63.0, 0.9995502677786328)
(64.0, 0.9995502677786328)
(65.0, 0.9995502677786328)
(66.0, 0.9995502677786328)
(67.0, 0.9995502677786328)
(68.0, 0.9995502677786328)
(69.0, 0.9995502677786328)
(70.0, 0.9995502677786328)
(71.0, 0.9995502677786328)
(72.0, 0.9995502677786328)
(73.0, 0.9995502677786328)
(74.0, 0.9995502677786328)
(75.0, 0.9995502677786328)
(76.0, 0.9995502677786328)
(77.0, 0.9995502677786328)
(78.0, 0.9995502677786328)
(79.0, 0.9995502677786328)
(80.0, 0.9995502677786328)
(81.0, 0.9995502677786328)
(82.0, 0.9995502677786328)
(83.0, 0.9995502677786328)
(84.0, 0.9995502677786328)
(85.0, 0.9995502677786328)
(86.0, 0.9995502677786328)
(87.0, 0.9995502677786328)
(88.0, 0.9995502677786328)
(89.0, 0.9995502677786328)
(90.0, 0.9995502677786328)
(91.0, 0.9995502677786328)
(92.0, 0.9995502677786328)
(93.0, 0.9995502677786328)
(94.0, 0.9995502677786328)
(95.0, 0.9995502677786328)
(96.0, 0.9995502677786328)
(97.0, 0.9995502677786328)
(98.0, 0.9995502677786328)
(99.0, 0.9995502677786328)
(100.0, 0.9995502677786328)
(101.0, 0.9995502677786328)
(102.0, 0.9995502677786328)
(103.0, 0.9995502677786328)
(104.0, 0.9995502677786328)
(105.0, 0.9995502677786328)
(106.0, 0.9995502677786328)
(107.0, 0.9995502677786328)
(108.0, 0.9995502677786328)
(109.0, 0.9995502677786328)
(110.0, 0.9995502677786328)
(111.0, 0.9995502677786328)
(112.0, 0.9995502677786328)
(113.0, 0.9995502677786328)
(114.0, 0.9995502677786328)
(115.0, 0.9995502677786328)
(116.0, 0.9995502677786328)
(117.0, 0.9995502677786328)
(118.0, 0.9995502677786328)
(119.0, 0.9995502677786328)
};
\addlegendentry{FOV}
\addplot+ [mark = {none}, thick, black, dash dot]coordinates {
(1.0, 0.0)
(2.0, 0.0)
(3.0, 0.0)
(4.0, 0.0)
(5.0, 0.0)
(6.0, 0.0)
(7.0, 0.0)
(8.0, 0.0)
(9.0, 0.0)
(10.0, 0.0)
(11.0, 0.0)
(12.0, 0.0)
(13.0, 0.0)
(14.0, 0.0)
(15.0, 0.0)
(16.0, 0.0)
(17.0, 0.0)
(18.0, 0.0)
(19.0, 0.0)
(20.0, 0.0)
(21.0, 0.0)
(22.0, 0.0)
(23.0, 0.0)
(24.0, 0.011630784752792372)
(25.0, 0.011630784752792372)
(26.0, 0.011630784752792372)
(27.0, 0.011630784752792372)
(28.0, 0.011630784752792372)
(29.0, 0.011630784752792372)
(30.0, 0.011630784752792372)
(31.0, 0.011630784752792372)
(32.0, 0.011630784752792372)
(33.0, 0.011630784752792372)
(34.0, 0.011630784752792372)
(35.0, 0.011630784752792372)
(36.0, 0.011630784752792372)
(37.0, 0.011630784752792372)
(38.0, 0.011630784752792372)
(39.0, 0.011630784752792372)
(40.0, 0.011630784752792372)
(41.0, 0.011630784752792372)
(42.0, 0.011630784752792372)
(43.0, 0.011630784752792372)
(44.0, 0.011630784752792372)
(45.0, 0.011630784752792372)
(46.0, 0.011630784752792372)
(47.0, 0.011630784752792372)
(48.0, 0.14644727924784168)
(49.0, 0.14644727924784168)
(50.0, 0.14644727924784168)
(51.0, 0.14644727924784168)
(52.0, 0.14644727924784168)
(53.0, 0.14644727924784168)
(54.0, 0.14644727924784168)
(55.0, 0.14644727924784168)
(56.0, 0.14644727924784168)
(57.0, 0.14644727924784168)
(58.0, 0.14644727924784168)
(59.0, 0.14644727924784168)
(60.0, 0.14644727924784168)
(61.0, 0.14644727924784168)
(62.0, 0.14644727924784168)
(63.0, 0.14644727924784168)
(64.0, 0.14644727924784168)
(65.0, 0.14644727924784168)
(66.0, 0.14644727924784168)
(67.0, 0.14644727924784168)
(68.0, 0.14644727924784168)
(69.0, 0.14644727924784168)
(70.0, 0.14644727924784168)
(71.0, 0.14644727924784168)
(72.0, 0.49530307508052784)
(73.0, 0.49530307508052784)
(74.0, 0.49530307508052784)
(75.0, 0.49530307508052784)
(76.0, 0.49530307508052784)
(77.0, 0.49530307508052784)
(78.0, 0.49530307508052784)
(79.0, 0.49530307508052784)
(80.0, 0.49530307508052784)
(81.0, 0.49530307508052784)
(82.0, 0.49530307508052784)
(83.0, 0.49530307508052784)
(84.0, 0.49530307508052784)
(85.0, 0.49530307508052784)
(86.0, 0.49530307508052784)
(87.0, 0.49530307508052784)
(88.0, 0.49530307508052784)
(89.0, 0.49530307508052784)
(90.0, 0.49530307508052784)
(91.0, 0.49530307508052784)
(92.0, 0.49530307508052784)
(93.0, 0.49530307508052784)
(94.0, 0.49530307508052784)
(95.0, 0.49530307508052784)
(96.0, 0.9999999225627546)
(97.0, 0.9999999225627546)
(98.0, 0.9999999225627546)
(99.0, 0.9999999225627546)
(100.0, 0.9999999225627546)
(101.0, 0.9999999225627546)
(102.0, 0.9999999225627546)
(103.0, 0.9999999225627546)
(104.0, 0.9999999225627546)
(105.0, 0.9999999225627546)
(106.0, 0.9999999225627546)
(107.0, 0.9999999225627546)
(108.0, 0.9999999225627546)
(109.0, 0.9999999225627546)
(110.0, 0.9999999225627546)
(111.0, 0.9999999225627546)
(112.0, 0.9999999225627546)
(113.0, 0.9999999225627546)
(114.0, 0.9999999225627546)
(115.0, 0.9999999225627546)
(116.0, 0.9999999225627546)
(117.0, 0.9999999225627546)
(118.0, 0.9999999225627546)
(119.0, 0.9999999225627546)
(120.0, 0.9999999225627546)
};
\addlegendentry{RFB}
\end{axis}

\end{tikzpicture}
	\caption{Evolution of belief uncertainty for instantaneous bearing, field-of-view, and rotate-for-bearing modalities during a single simulation.}
	\label{fig:snapshot}
\end{figure}
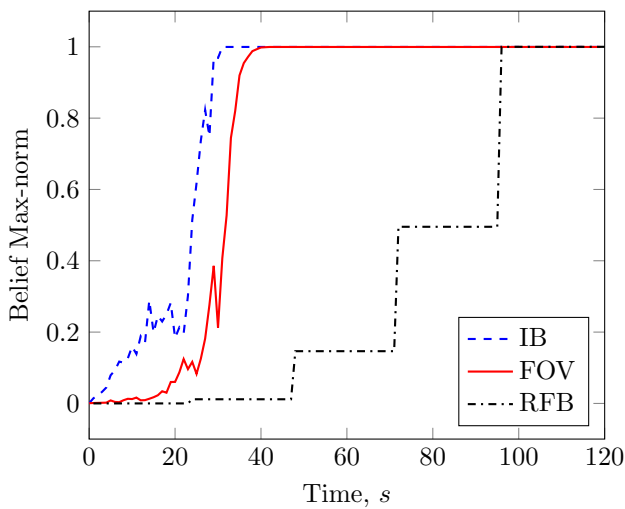

To better visualize how informative each measurement is, we run a single simulation and plot the belief max-norm with time.
The result can be seen in~\Cref{fig:snapshot}.
When using the IB and FOV methods, the belief max-norm rises much more quickly; this result satisfies intuition as measurements are made every second.
In contrast, the max-norm during the RFB method only jumps up every after each roation is completed.
While the FOV measurements are less informative than the IB measurements, the FOV performance is much closer to IB, despite being far easier to implement on a real UAV.

\subsection{Measurement Quality and Quantity}
We modeled measurement quality with the mistake rate $\mu$ and the cone width $\alpha$.
The latter results in uncertainty regions where either observation is equally likely.
We decided on a cone width of \ang{120} by examining experimentally derived gain patterns such as those shown in~\Cref{fig:two_432,fig:robust1}.
We selected a mistake rate of 0.1 because it intuitively seemed like a reasonable and conservative value.
However, the true rate might be lower.
Furthermore, better antennas might yield larger cone widths with smaller uncertainty regions, leading to faster localization.
It would be useful to know how antenna quality affects localization.

\begin{figure}
	\centering
	\begin{tikzpicture}[]
\begin{axis}[width = {3.3in}, xlabel = {Cone Width $\alpha$, degrees}, ylabel = {Localization Time, s}]\addplot+ coordinates {
(120.0, 30.065)
(140.0, 28.063)
(160.0, 26.361)
(180.0, 20.165)
};
\addlegendentry{$\mu=0.10$}
\addplot+ coordinates {
(120.0, 24.698)
(140.0, 24.034)
(160.0, 22.025)
(180.0, 16.919)
};
\addlegendentry{$\mu=0.05$}
\addplot+ [mark = {triangle*}, black,mark options={black}]coordinates {
(120.0, 21.435)
(140.0, 20.299)
(160.0, 19.438)
(180.0, 15.019)
};
\addlegendentry{$\mu=0.01$}
\end{axis}

\end{tikzpicture}
	\caption{As the cone width increases, the uncertainty region shrinks, leading to faster localization.}
	\label{fig:conewidth}
\end{figure}
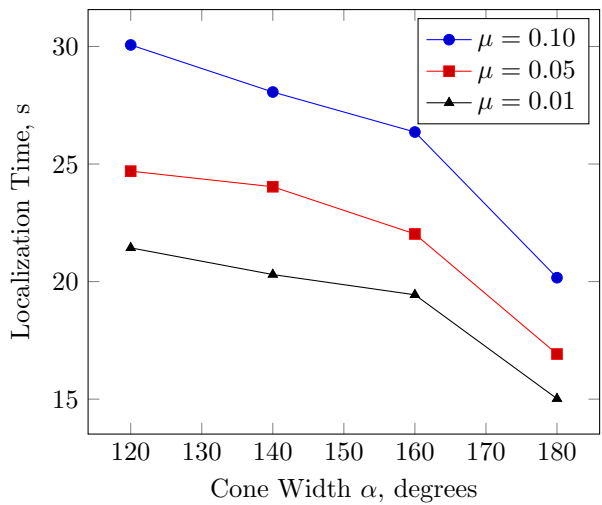

We ran 1000 simulations for different cone widths and mistake rates while using a greedy, information-theoretic policy.
Results can be seen in \Cref{fig:conewidth}.
As the cone width increases and the uncertainty region decreases, localization time decreases.
At the ideal cone width of \ang{180}, localization takes roughly two-thirds the time it does at a cone width of \ang{120}.
However, this reduction only corresponds to savings of \SI{10}{\second}, and the mechanical difficulty of building near-perfect antennas is probably not worth the time savings.
Likewise, a lower mistake rate $\mu$ reduces noise and localization time.

Localization depends not only on measurement quality but also measurement quantity.
Measurement quantity is dictated by the measurement sample rate.
In the RFB method, this rate is limited by the time to make a full rotation.
However, in the FOV method, the sample rate is limited only by the electronics involved.
Because the wildlife collar only transmits pulses at \SI{1}{\Hz}, we use that value as our sample rate across all transmitters.
However, a higher sample rate can be used if the transmitter of interest continuously transmits.
Higher sample rates yield more measurements which reduce localization time by providing more information about the transmitter's location.

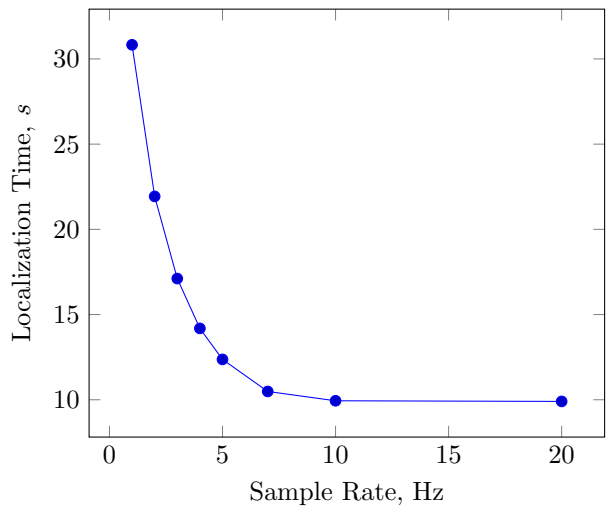
\begin{figure}
	\centering
	\begin{tikzpicture}[]
\begin{axis}[ylabel = {Localization Time, $s$}, xlabel = {Sample Rate, Hz}]\addplot+ coordinates {
(1.0, 30.829)
(2.0, 21.9325)
(3.0, 17.111666666666665)
(4.0, 14.18775)
(5.0, 12.368)
(7.0, 10.487285714285713)
(10.0, 9.9403)
(20.0, 9.90285)
};
\end{axis}

\end{tikzpicture}
	\caption{As the sample rate increases, the time to localization decreases.}
	\label{fig:rate}
\end{figure}

We ran 1000 simulations for different sample rates and using a greedy, information-theoretic policy.
The cone width and mistake rate were set to the defualt values of \ang{120} and 0.1.
Regardless of the sample rate, the UAV was limited to a planar speed of \SI{5}{\meter\per\second} and an angular speed of \SI{10}{\degree\per\second}.
The results can be seen in~\Cref{fig:rate}.
Increasing the sample rate can drastically improve localization time.
At \SI{10}{\Hz}, localization time is less than half the time to perform a single rotation in the RFB modality.

\section{Flight Tests}
\label{sec:experiments}
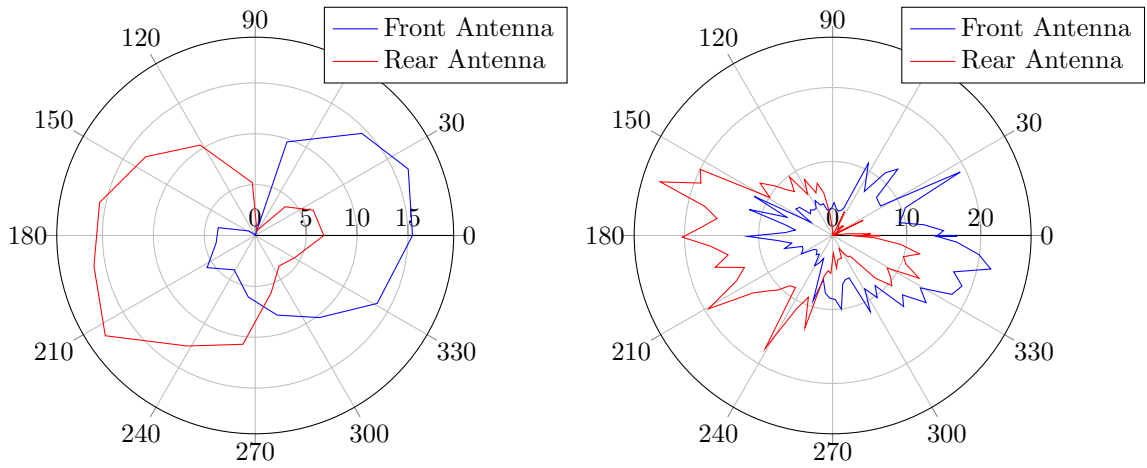
\begin{figure}
	\centering
	\begin{tikzpicture}[]
	\begin{polaraxis}[width=2.7in]\addplot+ [mark = {none}]coordinates {
(0.0, 15.447099999999999)
(23.48725889579785, 16.386799999999997)
(43.78256991492185, 14.504799999999996)
(71.26907422073285, 9.713999999999999)
(93.11740494365279, 0.0)
(121.2754157853571, 0.10229999999999961)
(144.2876926689915, 0.7646000000000015)
(167.99897806268547, 3.708199999999998)
(190.7310785845009, 3.906100000000002)
(213.78907750018672, 5.686799999999998)
(238.47250038487917, 3.9587999999999965)
(263.53573699196386, 6.080599999999997)
(285.3846803683646, 8.116700000000002)
(307.99239175285715, 10.2391)
(330.84332372138266, 13.714199999999998)
(0.0, 15.447099999999999)
};
\addlegendentry{Front Antenna}
\addplot+ [mark = {none}]coordinates {
(0.0, 6.737499999999997)
(23.48725889579785, 6.220199999999998)
(43.78256991492185, 4.0848999999999975)
(71.26907422073285, 0.5157999999999987)
(93.11740494365279, 5.195599999999999)
(121.2754157853571, 10.395600000000002)
(144.2876926689915, 13.2549)
(167.99897806268547, 15.624499999999998)
(190.7310785845009, 16.1252)
(213.78907750018672, 17.718899999999998)
(238.47250038487917, 12.746000000000002)
(263.53573699196386, 10.765999999999998)
(285.3846803683646, 5.866500000000002)
(307.99239175285715, 3.8046000000000006)
(330.84332372138266, 4.4589)
(0.0, 6.737499999999997)
};
\addlegendentry{Rear Antenna}
\end{polaraxis}

\end{tikzpicture}
	\begin{tikzpicture}[]
	\begin{polaraxis}[width=2.7in]\addplot+ [mark = {none}]coordinates {
(0.0, 13.668800000000001)
(359.6613819430777, 16.806700000000003)
(359.6751329301608, 14.067300000000003)
(2.535338243453907, 14.987800000000004)
(6.702460287440374, 12.409000000000006)
(10.911408250471396, 9.164500000000004)
(15.967760792500926, 9.8324)
(20.965671639427086, 10.528300000000002)
(26.46262872591221, 19.247200000000003)
(31.630708037992235, 7.622500000000002)
(36.19202504502873, 7.667999999999999)
(40.970493056419784, 7.851300000000002)
(45.54326921935889, 12.563100000000006)
(49.96191973540779, 11.0856)
(54.22873643574705, 7.325400000000002)
(58.854797673633314, 8.703800000000001)
(63.69228033792284, 10.859100000000005)
(67.94821084015457, 3.958600000000004)
(72.03454583502763, 3.5382999999999996)
(77.15335077672638, 3.411900000000003)
(82.56035348937597, 3.6996000000000038)
(87.37663671524568, 4.435500000000005)
(92.06717330517802, 3.711100000000002)
(96.86913258616943, 3.554400000000001)
(101.81776906271438, 3.5087000000000046)
(106.47591593712792, 4.479700000000001)
(111.39705043950659, 4.3992)
(116.69118046651539, 5.273800000000001)
(121.24275719103463, 4.286200000000001)
(126.32088212927914, 4.276400000000002)
(131.07012929311855, 4.5762)
(135.38564740604392, 4.1899000000000015)
(139.55276945003038, 5.8581)
(144.17080927878482, 6.115400000000001)
(148.77223333148044, 3.4391000000000034)
(153.4762168295045, 11.864600000000003)
(158.2753113215203, 6.698100000000004)
(162.70198324670102, 11.824300000000001)
(167.15042756809675, 6.6270000000000024)
(171.5994448472876, 5.0870999999999995)
(176.3767669430884, 6.371300000000005)
(180.6172275848516, 11.478200000000001)
(185.09603866938926, 7.549400000000006)
(190.54372138549311, 5.617900000000006)
(195.1606152986573, 6.7136)
(199.92590528076033, 4.360300000000002)
(205.2452454507549, 4.419700000000006)
(210.61672478010638, 4.899300000000004)
(215.7633754706485, 3.284000000000006)
(220.1105781496446, 2.936700000000002)
(224.72827420372198, 2.9646000000000043)
(229.546047119859, 3.422400000000003)
(234.433720887002, 3.0945000000000036)
(239.1913332346503, 4.727700000000006)
(243.80032033024168, 4.171900000000001)
(248.57065234094188, 3.3333000000000013)
(253.3366298723991, 9.190600000000003)
(258.2342158093979, 6.110200000000006)
(262.7735885328808, 7.854700000000001)
(267.2641509817966, 8.465600000000002)
(272.25143861824324, 8.661700000000003)
(276.80091889019013, 10.037100000000002)
(281.549644662436, 6.750700000000002)
(286.02771090183995, 6.1983999999999995)
(291.1866801849774, 6.148300000000006)
(296.2838273220202, 11.4775)
(301.0767911656281, 8.4116)
(305.6197162274409, 10.336500000000001)
(310.2342037178655, 8.871200000000002)
(314.9494744844536, 13.572900000000004)
(319.6316282904832, 12.385400000000004)
(324.26187212027395, 15.550000000000004)
(329.0906458260775, 13.787800000000004)
(333.7098315704222, 17.963500000000003)
(338.6601869203525, 18.754600000000003)
(343.27249717115563, 17.0773)
(348.01716067263396, 21.827700000000004)
(352.5292033092892, 19.931600000000003)
(356.91347635763026, 16.798600000000004)
(0.0, 13.668800000000001)
};
\addlegendentry{Front Antenna}
\addplot+ [mark = {none}]coordinates {
(0.0, 2.5321)
(359.6613819430777, 6.342700000000001)
(359.6751329301608, 3.764700000000005)
(2.535338243453907, 5.1544000000000025)
(6.702460287440374, 1.9923000000000002)
(10.911408250471396, 0.44819999999999993)
(15.967760792500926, 0.7797000000000054)
(20.965671639427086, 0.2698999999999998)
(26.46262872591221, 4.609100000000005)
(31.630708037992235, 0.8854000000000042)
(36.19202504502873, 0.4183000000000021)
(40.970493056419784, 1.9236000000000004)
(45.54326921935889, 1.867200000000004)
(49.96191973540779, 1.0268000000000015)
(54.22873643574705, 1.3781000000000034)
(58.854797673633314, 1.494800000000005)
(63.69228033792284, 3.5830000000000055)
(67.94821084015457, 0.0)
(72.03454583502763, 0.13589999999999947)
(77.15335077672638, 0.12190000000000367)
(82.56035348937597, 0.3132000000000019)
(87.37663671524568, 0.3464000000000027)
(92.06717330517802, 0.7173000000000016)
(96.86913258616943, 2.8885000000000005)
(101.81776906271438, 5.894300000000001)
(106.47591593712792, 7.348400000000005)
(111.39705043950659, 6.1997)
(116.69118046651539, 8.460700000000003)
(121.24275719103463, 6.546600000000005)
(126.32088212927914, 9.789300000000004)
(131.07012929311855, 7.878900000000002)
(135.38564740604392, 8.8493)
(139.55276945003038, 10.362500000000004)
(144.17080927878482, 12.101200000000006)
(148.77223333148044, 9.915300000000002)
(153.4762168295045, 19.988000000000003)
(158.2753113215203, 19.1459)
(162.70198324670102, 24.363100000000003)
(167.15042756809675, 17.723100000000002)
(171.5994448472876, 15.785300000000003)
(176.3767669430884, 17.7434)
(180.6172275848516, 20.289400000000004)
(185.09603866938926, 16.5103)
(190.54372138549311, 14.354900000000004)
(195.1606152986573, 16.4186)
(199.92590528076033, 12.6736)
(205.2452454507549, 14.347600000000003)
(210.61672478010638, 19.471600000000002)
(215.7633754706485, 13.3168)
(220.1105781496446, 11.660900000000005)
(224.72827420372198, 10.477200000000003)
(229.546047119859, 8.936)
(234.433720887002, 8.625300000000003)
(239.1913332346503, 17.7313)
(243.80032033024168, 11.028000000000006)
(248.57065234094188, 8.9784)
(253.3366298723991, 13.009300000000003)
(258.2342158093979, 5.599600000000002)
(262.7735885328808, 4.7989999999999995)
(267.2641509817966, 5.125399999999999)
(272.25143861824324, 2.415600000000005)
(276.80091889019013, 4.447700000000005)
(281.549644662436, 3.3745000000000047)
(286.02771090183995, 3.2824000000000026)
(291.1866801849774, 3.290200000000006)
(296.2838273220202, 1.9928000000000026)
(301.0767911656281, 2.7105000000000032)
(305.6197162274409, 3.5307999999999993)
(310.2342037178655, 3.6410000000000053)
(314.9494744844536, 8.136200000000002)
(319.6316282904832, 10.482400000000005)
(324.26187212027395, 9.6466)
(329.0906458260775, 8.588000000000001)
(333.7098315704222, 13.014500000000005)
(338.6601869203525, 10.629800000000003)
(343.27249717115563, 9.982)
(348.01716067263396, 11.9846)
(352.5292033092892, 9.274300000000004)
(356.91347635763026, 5.676300000000005)
(0.0, 2.5321)
};
\addlegendentry{Rear Antenna}
\end{polaraxis}

\end{tikzpicture}
	\caption{(Left) Signal strength measurements made \SI{20}{\meter} from our wildlife collar. (Right) Signal strength measurements made \SI{100}{\meter} from a cell phone placing a voice call over LTE.}
	\label{fig:collarphone}
\end{figure}

To validate our sensing modality across different transmitters and frequencies, we rotated the UAV in place and measured the strength received by each antenna as the relative bearing to the transmitter changed.
The results for the UV-5R radio were shown in \Cref{fig:two_432,fig:robust1}.
The results for the wildlife collar and cell phone are shown in \Cref{fig:collarphone}.
In all cases, the front-facing antenna receives higher strength when the UAV faces the RF source; the rear receives higher strength when the UAV is facing away.
Note that the cell phone's transmitted power changes rapidly with time, but these changes affect both antennas equally, validating our claims of robustness to time-varying effects.

\begin{figure*}
	\centering
	\includegraphics[width=2.1in,trim={1.05in 0.33in 1.05in 0.35in},clip]{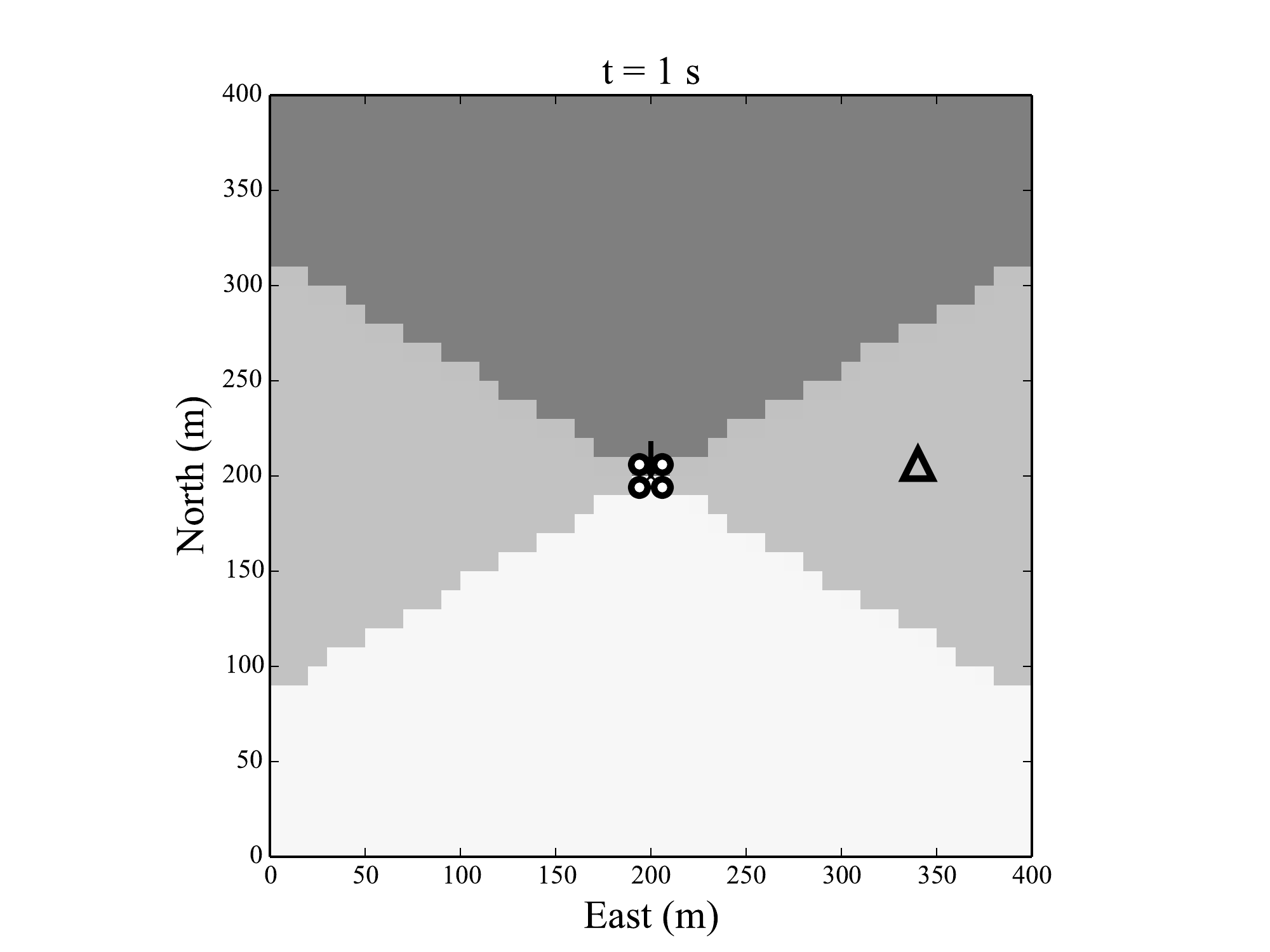}
	\includegraphics[width=2.1in,trim={1.05in 0.33in 1.05in 0.35in},clip]{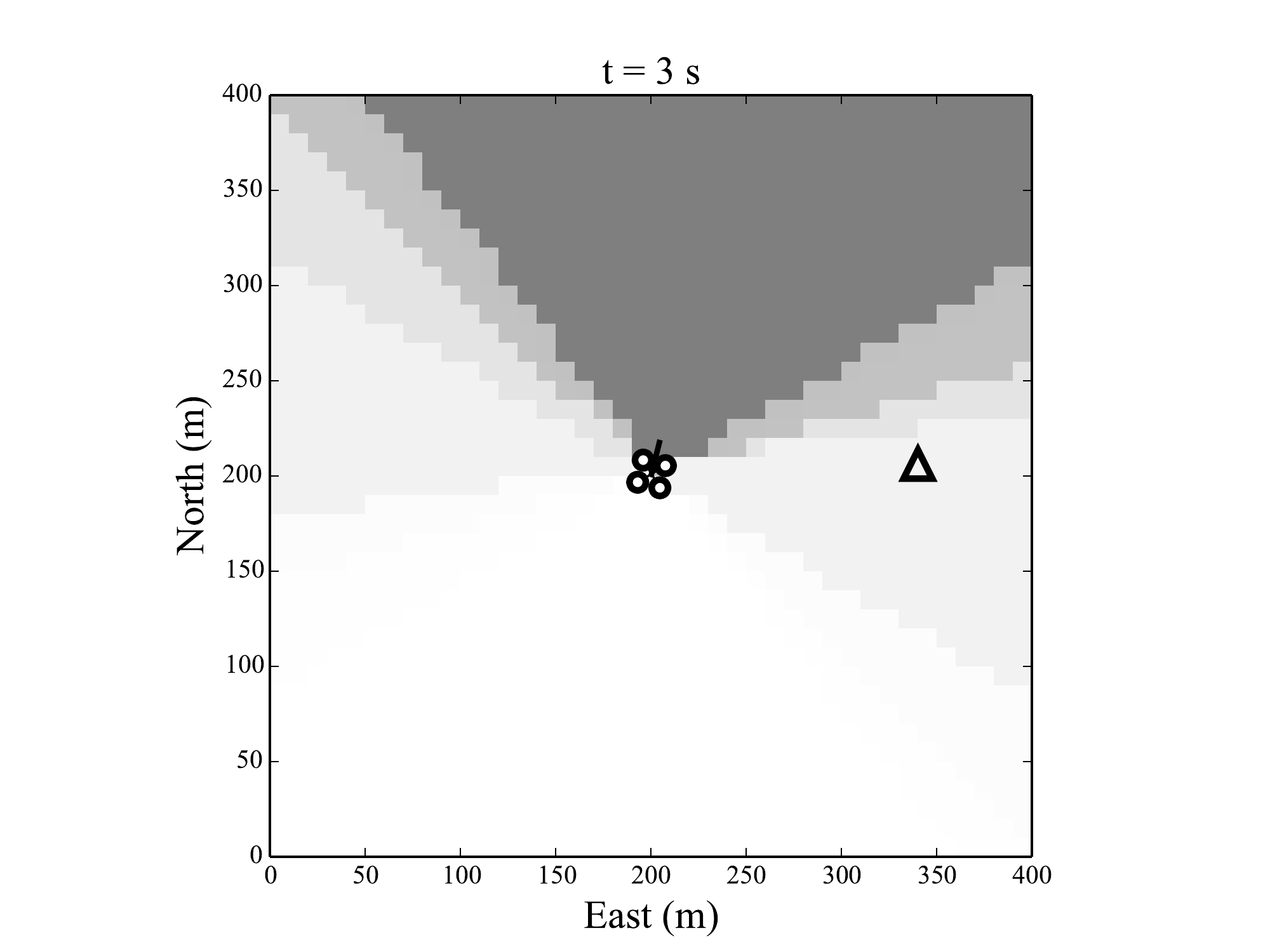}
	\includegraphics[width=2.1in,trim={1.05in 0.33in 1.05in 0.35in},clip]{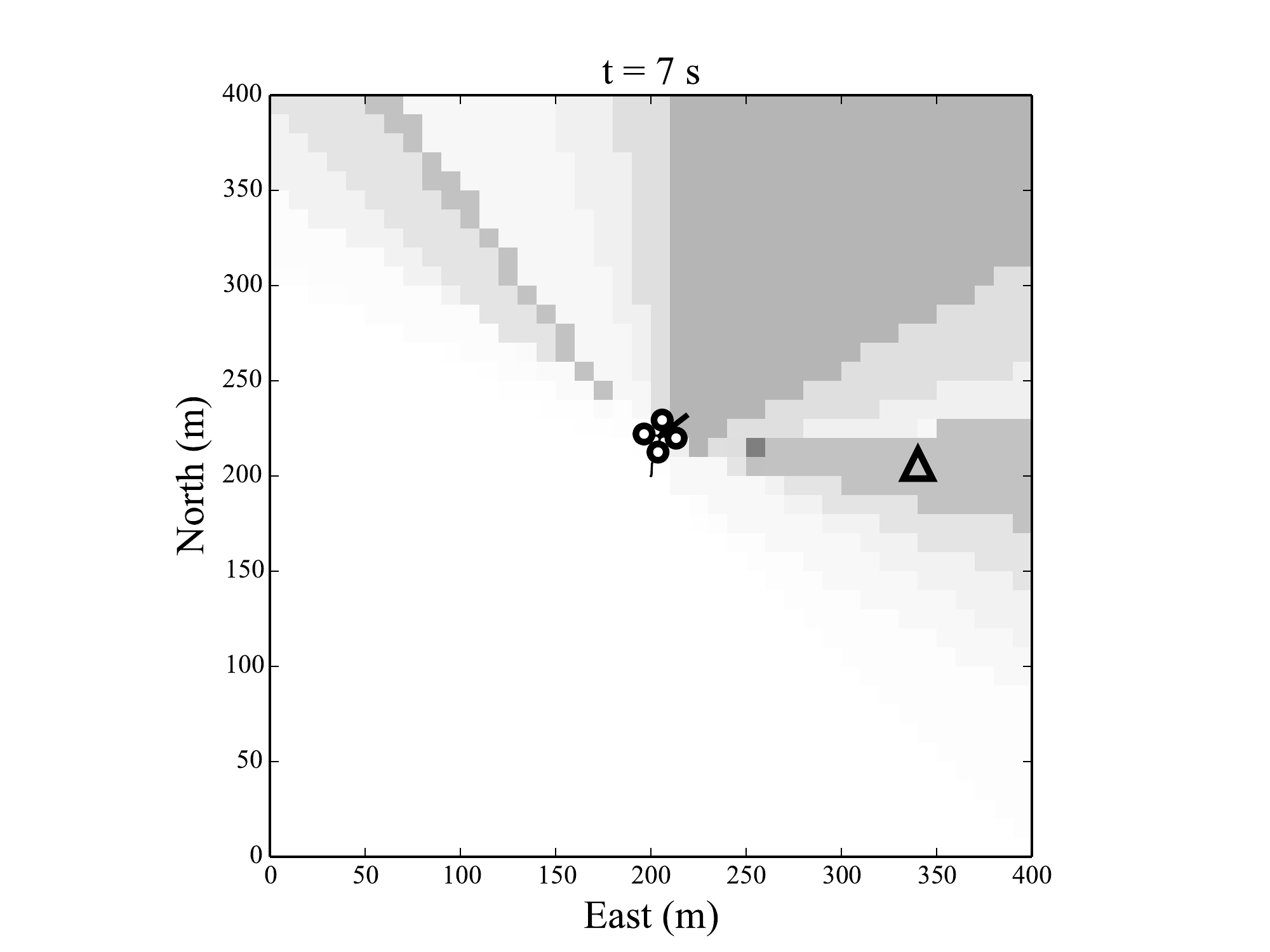}
	\includegraphics[width=2.1in,trim={1.05in 0 1.05in 0.35in},clip]{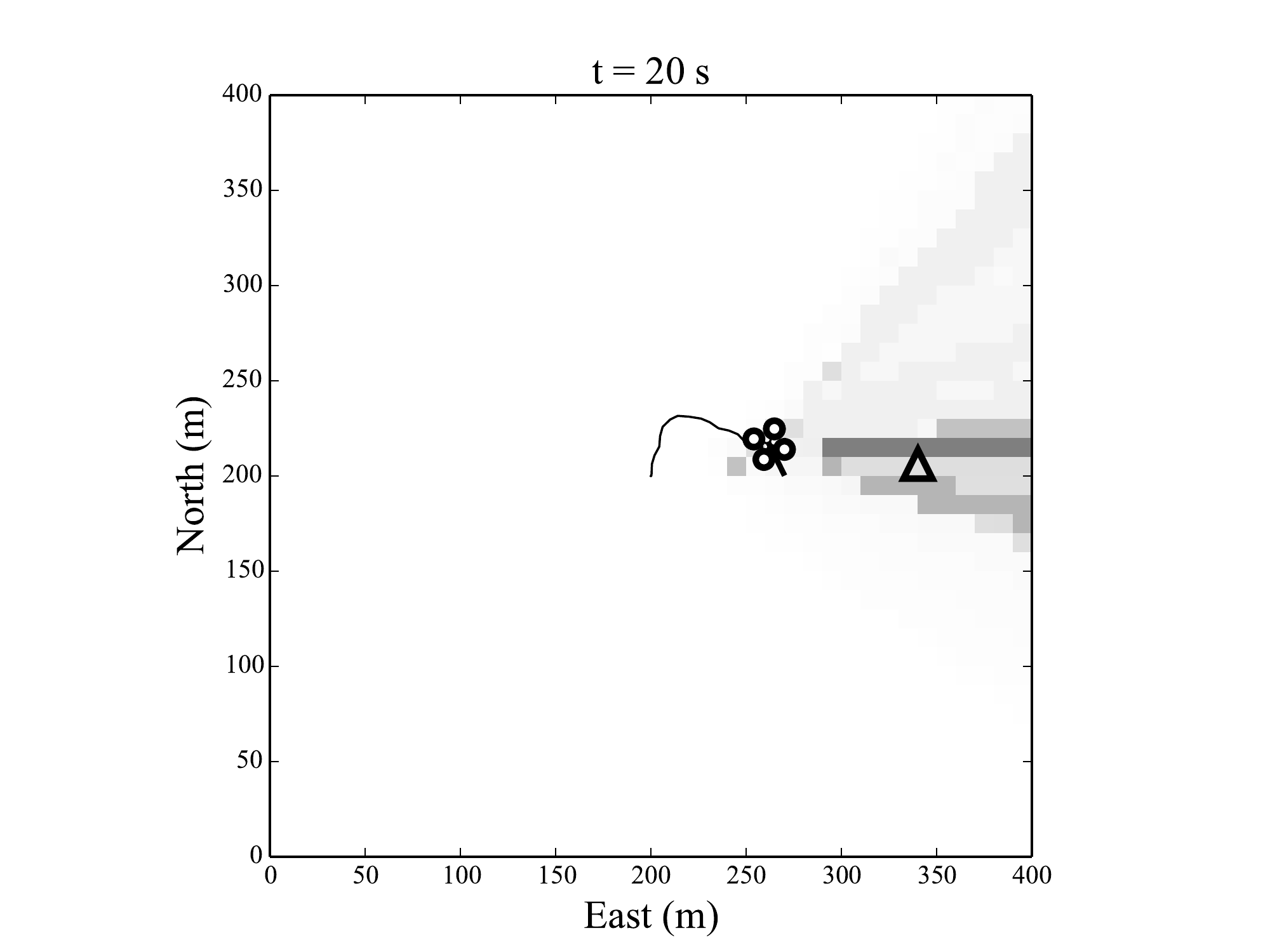}
	\includegraphics[width=2.1in,trim={1.05in 0 1.05in 0.35in},clip]{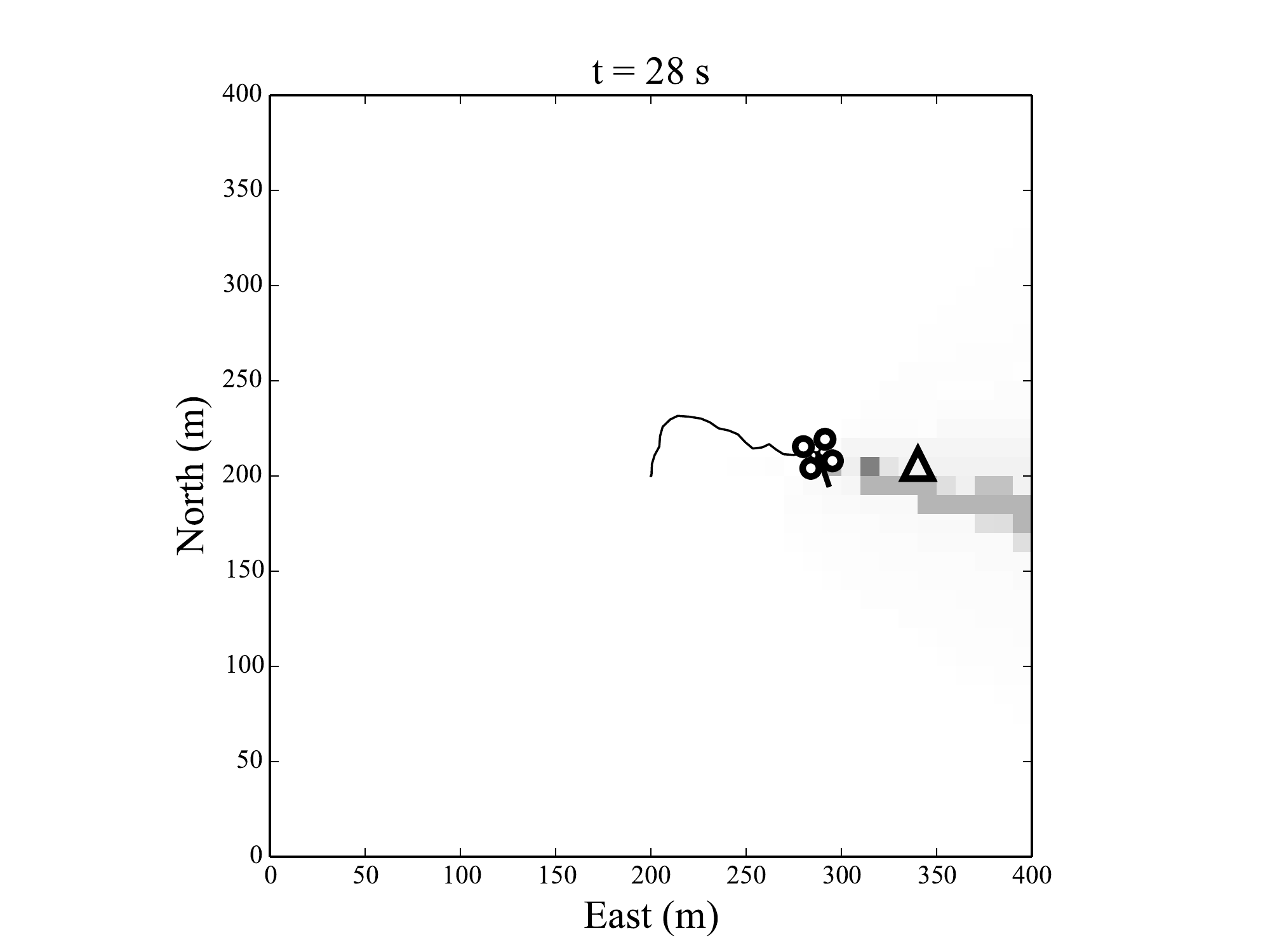}
	\includegraphics[width=2.1in,trim={1.05in 0 1.05in 0.35in},clip]{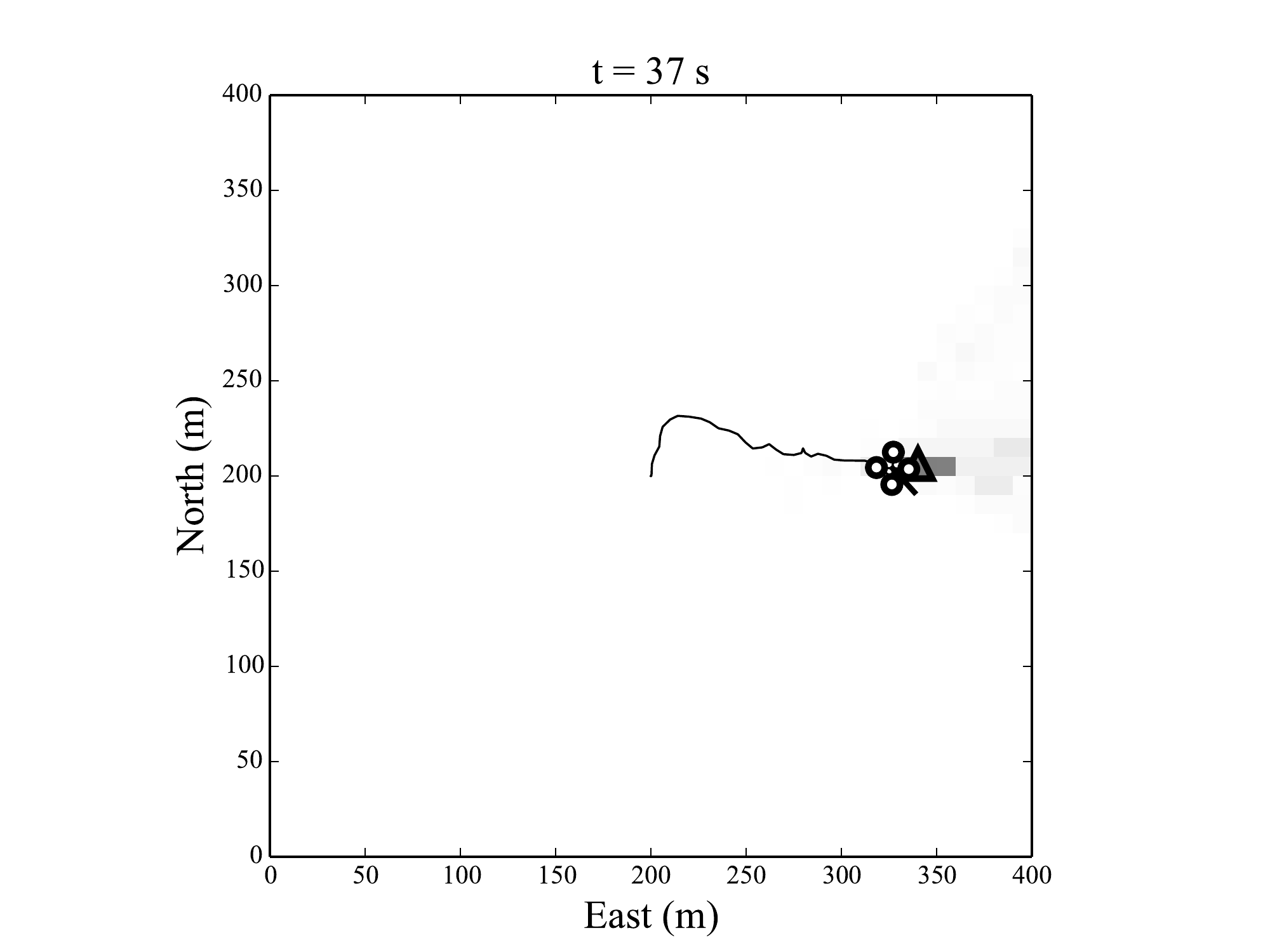}
	\caption{Flight test trajectory using UV-5R radio (triangle). After 37 seconds, the UAV is fairly certain of the radio's location.}
	\label{fig:localization}
\end{figure*}

We then flew autonomous localization tests in a $\SI{400}{\meter} \times \SI{400}{\meter}$ search area with the different transmitters.
The UAV flew at an altitude of \SI{10}{\meter}, moved at \SI{5}{\meter\per\second}, made measurements at \SI{1}{\Hz}, and used a greedy, information-theoretic policy.
\Cref{fig:localization} shows one flight test trajectory.
The beliefs shown were generated on the UAV; no post-processing was done.
After \SI{37}{\second}, the UAV is fairly certain of the radio's location, which is in line with simulation results---localization occurs in roughly the time it would take to perform one rotation.

Using the received observations and the GPS coordinates of the UAV and transmitter, we can estimate the true mistake rate and validate our model.
Across the localization attempts, the UAV made 179 measurements where the transmitter was either in the front or rear cone; of these, 166 observations corresponded to the correct cone.
This ratio corresponds to a mistake rate of 0.073, which is slightly less than the value of 0.1 we assumed earlier.
This mistake rate held across the different transmitters; even though the phone's strength varied rapidly with time, it only made 4 mistakes in 79 observations, again validating our claim of robustness.
Of the 203 measurements made when a transmitter was in the uncertainty region between the cones, a measurement of 1 was observed 111 times, or 54.7\% of the time. 
This value agrees with our model, which assume either observation was equally likely in the uncertainty region.


\section{Conclusion}
We present a system that significantly advances UAV-based RF localization in several key areas: in performance, where we localize RF sources in the time previous methods made a single measurement; in applicability, where we demonstrate our system across multiple frequencies and emission types; in robustness, where we continue to localize despite time-varying strength effects; and in cost, where our RF subsystem (dongles and antennas) is under \$50 USD, comparable to the cost of a single COTS directional antenna.
The end result is a versatile RF localizer that can be leveraged as a testbed for algorithmic development.

Future work will focus on control algorithms and reducing vulnerabilities of the sensing modality.
Flight tests showed the system dynamics are stochastic, and simulations showed the great advantage of greedy control---perhaps non-myopic control will perform even better.
We can both model transition stochasticity and plan non-myopically with the partially observable Markov decision process framework\cite{hsu2008point,Dressel2017}.
We also plan to expand the model to track non-stationary RF sources.
We can also expand the model to handle intermittent or weak transmissions.
Our sensing modality is vulnerable to the presence of multiple RF sources at the same frequency, as the model assumes a single transmitter.
However, this problem is not unique to our modality, as rotate-for-bearing schemes would also struggle in the presence of multiple transmitters.
Thus, the problem of hunting multiple transmitters with a UAV merits further investigation.

%

\section*{Acknowledgments}

We are grateful to DJI for donating the M-100 and Manifold.
This work was supported by NSF grant DGE-114747.

\bibliographystyle{aiaa}
\bibliography{references}

\end{document}